\pdfoutput=1
\documentclass{article}

\usepackage[preprint]{neurips_2026}

\usepackage[utf8]{inputenc}
\usepackage[T1]{fontenc}
\usepackage[colorlinks=true,linkcolor={blue!50!black},citecolor={blue!50!black},urlcolor={blue!50!black}]{hyperref}
\usepackage{url}
\usepackage{booktabs}
\usepackage{array}
\usepackage{graphicx}
\usepackage{amsfonts}
\usepackage{nicefrac}
\usepackage{microtype}
\usepackage{xcolor}
\usepackage{amsmath,amssymb,amsthm,mathtools}
\usepackage{tikz}
\usetikzlibrary{arrows.meta,positioning}

\newtheorem{theorem}{Theorem}
\newtheorem{proposition}[theorem]{Proposition}
\newtheorem{corollary}[theorem]{Corollary}
\newtheorem{lemma}[theorem]{Lemma}
\theoremstyle{definition}

\theoremstyle{remark}
\newtheorem{remark}[theorem]{Remark}

\newcommand{\E}{\mathbb{E}}
\newcommand{\Prob}{\mathbb{P}}
\newcommand{\Var}{\mathrm{Var}}
\newcommand{\KL}{\mathrm{KL}}
\newcommand{\kl}{\mathrm{kl}}
\newcommand{\TV}{\mathrm{TV}}
\newcommand{\cS}{\mathcal{S}}
\newcommand{\cA}{\mathcal{A}}
\newcommand{\cZ}{\mathcal{Z}}
\newcommand{\cR}{\mathcal{R}}
\newcommand{\pref}{\pi_{\mathrm{ref}}}
\newcommand{\dref}{d_{\mathrm{ref}}}
\newcommand{\SubOpt}{\mathrm{SubOpt}}
\newcommand{\cu}{c_{\mathrm u}}
\newcommand{\dagg}{\dagger}

\newif\ifneuripschecklist
\neuripschecklistfalse 

\newcommand{\CFivePsviSmallNMax}{0.41}

\newcommand{\CNineteenPooledSEPessimistic}{0.6}
\newcommand{\CNineteenPooledSEPlugin}{14}
\newcommand{\CNineteenPooledSlopeBern}{2.16}
\newcommand{\CNineteenPooledSlopeHoe}{3.06}

\newcommand{\CTwentyAdditiveMaxUnclipped}{1.61}
\newcommand{\CTwentyFourCtrl}{261}
\newcommand{\CTwentyOneHCross}{222}

\newcommand{\CTwentyOneLoopNRefMin}{3.6\cdot10^{11}}
\newcommand{\CTwentyOneLoopRatioMin}{3.4\cdot10^{4}}

\newcommand{\CTwentyOneLoopTOneNRefMin}{6.6\cdot10^{9}}
\newcommand{\CTwentyOnePlugNRef}{3.2\cdot10^{7}}
\newcommand{\CTwentyOnePlugRatioMin}{11.5}

\newcommand{\TotalCountedMinutes}{19}

\hypersetup{pdftitle={Offline KL-Regularized Reinforcement Learning with Stochastic Transitions},pdfauthor={Andrew Liu}}

\title{Offline KL-Regularized Reinforcement Learning\\with Stochastic Transitions}

\author{Andrew Liu\\
Columbia University\\
\texttt{all2209@columbia.edu}}

\begin{document}

\maketitle

\begin{abstract}
KL-regularized reinforcement learning is the objective behind RL fine-tuning of language models, and its data are often collected by
the reference policy itself. However, beyond bandits and deterministic transitions, little is known about how much offline data it
needs. In this work, we study finite discounted Markov decision processes with unknown stochastic transitions, $K$ state-action pairs and
effective horizon $H$. For data that cover the regularized optimum with concentrability $C\ge4$, every estimator needs
$\Omega(KCH^3/(\epsilon\max\{\epsilon,\beta\}))$ samples on some instances when $\epsilon$ is small, where $\beta$ is the temperature. Under a fixed
design, pessimistic soft value iteration with a variance-aware penalty attains this rate with high probability, up to logarithmic factors,
when $\beta\ge1.4H$. When the data
come from the reference policy itself, we show that it covers the optimum up to a power law whose exponent grows as $H/\beta$; in the same
temperature range, this power law alone gives the estimator a sample complexity polynomial in $1/\epsilon$, with no coverage assumption. For
$\beta\ge H$ the power can also increase the sample size needed by a factor polynomial in $1/\epsilon$. Finally, the plug-in estimator, which is
minimax optimal with a generative model, can fail on reference data at small temperatures when many actions are rarely seen, while the
pessimistic estimator does not.
\end{abstract}

\section{Introduction}\label{sec:intro}

KL-regularized reinforcement learning (RL) maximizes expected reward minus $\beta$ times the KL divergence from a reference
policy $\pref$. It is the objective of RL fine-tuning of language models \citep{ouyang2022training} and, with a uniform
reference and the same number of actions at every state, of maximum-entropy RL \citep{ziebart2010modeling,haarnoja2018soft}. In the offline version of the problem the learner
sees a fixed dataset, often collected by $\pref$ itself, and must return a policy whose regularized value is close to that
of the regularized optimum $\pi^*$. We ask how much data this takes in a finite discounted Markov decision process (MDP) with
unknown stochastic transitions, as a function of the number of state-action pairs $K$, the effective horizon
$H=1/(1-\gamma)$, the temperature $\beta$ and the coverage of the data.

Two features separate this problem from offline KL-regularized multi-armed bandits with reference data, where the minimax rate is known \citep{ji2026optimal},
after earlier upper and lower bounds \citep{zhao2025sharp,zhao2026towards}. The first is the horizon. With a generative model the plug-in estimator
needs $\widetilde O(KH^3/(\epsilon\max\{\epsilon,\beta\}))$ samples, and no algorithm needs fewer \citep{liu2026sample}. For offline
data under single-policy coverage, the best previous upper bound for tabular KL-regularized MDPs is of order $H^7SAC^*/(\beta\epsilon)$ transitions in the
finite-horizon model \citep{hao2026fast}, and the lower bounds for this objective that apply to offline data either have horizon one \citep{zhao2025sharp,zhao2026towards,ji2026optimal}
or come from the generative model, where the coverage of the data plays no role. The second is the coverage of the
reference policy. With deterministic transitions and a fixed horizon, as in token-level models of text generation,
$\pi^*$ gives every trajectory at most $e^{H/\beta}$ times its reference probability, so data from $\pref$ cover $\pi^*$ up to that
factor \citep{foster2025good}. With stochastic transitions this argument fails; the bound of \citet{hao2026fast} multiplies per-step likelihood ratios and
grows as $e^{H(H+1)/\beta}$.

In this work, we give lower and upper bounds for both kinds of data (Table~\ref{tab:overview}). When the data cover $\pi^*$ with
concentrability $C$, every estimator needs $KCH^3/(\epsilon\max\{\epsilon,\beta\})$ samples on some instances (Theorem~\ref{thm:lb-design}), and
pessimistic soft value iteration (PSVI) with a floored Bernstein penalty matches this up to logarithmic factors when $\beta\ge1.4H$
(Theorem~\ref{thm:h3}). For every temperature it attains an $H^4$ rate (Theorem~\ref{thm:h4}), and $H^3$ after a burn-in
(Proposition~\ref{prop:burnin}). The analysis rests on a path-space form of pessimism (Lemma~\ref{lem:pathpess}) and, for $\beta\ge1.4H$, on an
energy bound for the variance of the estimation error along the optimum's path. Data collected by $\pref$ cover $\pi^*$ up to a power law, in the sense that a path event of probability
$p$ under $\pi^*$ has probability at least $e^{-c_1H/\beta}p^{1+c_2H/\beta}$ under $\pref$ (Theorem~\ref{thm:coverage}). A power of order $H/\beta$ is
necessary on chains (Lemma~\ref{lem:chain}) and can multiply the sample size by a factor polynomial in $1/\epsilon$ (Theorem~\ref{thm:power}), yet the
coverage bound alone is enough for PSVI to learn from reference data with no coverage assumption (Corollary~\ref{cor:refdata}). Finally, we show that on
reference data the plug-in estimator can lose a constant fraction of $H$ on instances with constant coverage and about $n/H$ actions at one state
when $\beta\lesssim H/\log n$ (Theorem~\ref{thm:plugin}), while PSVI does not (Section~\ref{sec:plugin}).

\begin{table}[t]
\caption{Sample complexities up to logarithmic factors and absolute constants, for $\epsilon\lesssim\min\{H,H^2/\beta\}$; $C$ is the concentrability of the
data with respect to $\pi^*$. Lower bounds are for the expected loss, on families of instances with $C\ge4$ (and enough pairs for
Theorem~\ref{thm:power}); upper bounds hold with high probability. Under (D2), $C=C^*(\pref)$; Theorem~\ref{thm:power} has no factor $C$, and its
second factor is the excess over the fixed-design lower bound at constant $C$.}
\label{tab:overview}
\centering
\small
\begin{tabular}{@{}>{\raggedright\arraybackslash}p{2.0cm}>{\raggedright\arraybackslash}p{4.9cm}>{\raggedright\arraybackslash}p{6.2cm}@{}}
\toprule
Data & Lower bound & Upper bound (PSVI) \\
\midrule
fixed design (D1) & $KCH^3/(\epsilon\max\{\epsilon,\beta\})$ (Theorem~\ref{thm:lb-design}) & $KCH^3/(\beta\epsilon)$ for $\beta\ge1.4H$ (Theorem~\ref{thm:h3});
$\min\{KCH^4/\epsilon^2,KC(H^4/\beta+H^2)/\epsilon\}$ for every $\beta$ (Theorem~\ref{thm:h4}); $KCH^3/(\beta\epsilon)$ for every $\beta$ after a burn-in
$n\mu\gtrsim H\iota L_n$ on the support of $d^*$ (Proposition~\ref{prop:burnin}) \\
reference trajectories (D2) & $KCH^3/(\epsilon\max\{\epsilon,\beta\})$ for $C\le2+2e^{\gamma^2H/(3\beta)}$ (Theorem~\ref{thm:gate}); for $\beta\ge H$,
$\frac{KH^3}{\beta\epsilon}\big(\frac{c'H^2}{\beta\epsilon}\big)^{\kappa H/\beta}$ (Theorem~\ref{thm:power}) & as for (D1) without the burn-in result, on pairs with
expected count at least of order $H\iota$; the other pairs can cost a factor $H$ (Theorems~\ref{thm:h4}(b) and~\ref{thm:h3}(b)); with no
coverage assumption, $H\big(\frac{KH^3}{\beta\epsilon}\big)^{1+c_2H/\beta}$ for $\beta\ge1.4H$ (Corollary~\ref{cor:refdata}) \\
\bottomrule
\end{tabular}
\end{table}

\section{Related work}\label{sec:related}

\paragraph{Offline KL-regularized learning with horizon one.} The closest results are for bandits. \citet{ji2026optimal} settle the
offline rate for multi-armed bandits with data from the reference policy, for both $\epsilon\le\beta$ and $\epsilon>\beta$; in our notation it reads
$KC^*/(\epsilon\max\{\epsilon,\beta\})$ at $H=1$. For $\beta\ge1.4H$, Theorems~\ref{thm:h3} and~\ref{thm:lb-design} extend its part $\epsilon\le\beta$ to MDPs with a
fixed design, with an extra factor $H^3$. In contextual bandits, \citet{zhao2025sharp} prove a lower bound that applies to offline algorithms
and a fast rate with a two-stage sampling scheme, and \citet{zhao2026towards} analyze pessimistic estimators under a coverage coefficient
that our $C_\Sigma$ extends to MDPs. \citet{xiong2023iterative} study the same bandit model of RLHF offline, online and in a hybrid setting.

\paragraph{Offline KL-regularized MDPs.} \citet{hao2026fast} give fast rates for tabular finite-horizon models: a pessimism-free
Q-learning with a factor $e^{H(H+1)/\beta}$, and a pessimistic one with $\widetilde O(H^7SAC^*/(\beta\epsilon))$ transitions, without lower bounds.
Other work obtains slow rates with function approximation \citep{uehara2023offline,liu2026complexity}, a rate in the least stationary
visitation probability of a behavior policy for policy mirror descent \citep{chen2026fast}, guarantees for off-policy policy gradient
\citep{cao2026anatomy}, and offline learning of KL-regularized games without pessimism \citep{zhang2026beyond,chen2026offline}. The analysis of
\citet{chen2026offline} for Markov games multiplies per-step likelihood ratios, which Theorem~\ref{thm:coverage} avoids by a moment bound on the
whole path.

\paragraph{Unregularized offline RL.} Pessimism is efficient under single-policy concentrability
\citep{jin2021pessimism,rashidinejad2021bridging,xie2021policy}. For tabular discounted MDPs with independent samples, value iteration with a
floored Bernstein penalty attains the minimax rate $H^3SC^*/\epsilon^2$ without sample splitting, and for episodic trajectories a two-fold
subsampling of the trajectories gives the minimax rate \citep{li2024settling}; pessimistic Q-learning is near-optimal as well
\citep{shi2022pessimistic}. PSVI combines the floored penalty of the first with the subsampling of the second. In batch bandits every
confidence-adjusted index rule, including the greedy one, is minimax optimal over all instances \citep{xiao2021optimality}; Theorem~\ref{thm:plugin}
concerns instances with constant single-policy coverage instead. Regularization makes the loss
a KL divergence between path laws (Lemma~\ref{lem:pathkl}), which gives $1/\epsilon$ rates through one path sum (Lemma~\ref{lem:pathpess}), and gives
$\pi^*$ full support, so every action at every visited state must be covered.

\paragraph{Regularized MDPs with a generative model or online exploration.} \citet{azar2013minimax} prove the minimax $H^3$ rate for the
unregularized problem, and \citet{kozuno2022kl} reach it with KL and entropy regularization used as a tool. For regularized best-policy
identification with online exploration, \citet{tiapkin2023fast,tiapkin2024demonstration} obtain rates of order $1/(\lambda\epsilon)$ in the
regularization weight $\lambda$, and \citet{tiapkin2023fast} conjecture $H^2SA/(\lambda\epsilon)$ episodes under entropy regularization.

\paragraph{Coverage of the reference policy.} Theorem~\ref{thm:coverage} extends the deterministic-transition bound of \citet{foster2025good}
and \citet{xie2024exploratory} (Section~\ref{sec:intro}) to stochastic transitions. A learner limited to fresh rollouts of the reference and a constant number of reward queries can need exponentially
many generator queries in $H$ \citep{rege2026role}. \citet{song2024importance} give guarantees for offline KL-regularized fine-tuning on reference data under a
coverage condition and conjecture that the plug-in fails under single-policy coverage; the instance of Theorem~\ref{thm:plugin} makes a
single decision, so it gives a failure of this kind for observed rewards rather than preferences. \citet{huang2024correcting} argue that KL
regularization does not prevent overoptimization under single-policy coverage and replace it by a $\chi^2$-type regularizer. \citet{wu2025greedy} show that greedy
estimation suffices for KL-regularized contextual bandits from preference data, at the cost of a factor exponential in the reward range
over $\beta$.

\section{Preliminaries}\label{sec:setting}

\paragraph{Regularized MDPs.} A finite discounted MDP $\mathcal M=(\cS,\{\cA_s\}_{s\in\cS},P,r,\gamma)$ has state-action pairs
$\cZ=\{(s,a):s\in\cS,\,a\in\cA_s\}$, $K=|\cZ|$, discount $\gamma\in[1/2,1)$ and effective horizon $H=1/(1-\gamma)$. The mean reward
$r:\cZ\to[0,1]$ and the transition kernel $P$ are unknown. A reference policy $\pref$ with $\pref(a|s)>0$ for every $a\in\cA_s$ and a
temperature $\beta>0$ define the regularized value of a policy $\pi$,
\[
V^\pi(s)=\E^\pi_s\Big[\sum_{t\ge0}\gamma^t\big(r(s_t,a_t)-\beta\KL\big(\pi(\cdot|s_t)\,\|\,\pref(\cdot|s_t)\big)\big)\Big],
\]
where a history-dependent policy pays $\beta\KL(\pi(\cdot|h_t)\|\pref(\cdot|s_t))$ at step $t$. The soft-optimal values satisfy
$Q^*=r+\gamma PV^*$ and $V^*(s)=\beta\log\sum_a\pref(a|s)e^{Q^*(s,a)/\beta}$, and the regularized optimum is
$\pi^*(a|s)=\pref(a|s)\exp((Q^*(s,a)-V^*(s))/\beta)$. Since $r\in[0,1]$, $0\le Q^*\le H$ and $0\le V^*\le H$. For an initial
distribution $\rho$ on $\cS$, the suboptimality of $\pi$ is $\SubOpt_\rho(\pi)=V^*(\rho)-V^\pi(\rho)$.

\paragraph{Killed chains and occupancies.} We represent discounting by killing. From $s_0\sim\rho$, at each step $t$ the
chain draws $a_t\sim\pi(\cdot|s_t)$; then, with probability $1-\gamma$ and independently of everything else, it stops and sets
$T=t$; otherwise it moves to $s_{t+1}\sim P(\cdot|s_t,a_t)$. We write $z_t=(s_t,a_t)$, $\Prob^\pi$ for the law of the killed path
$(z_0,\dots,z_T)$ under $\pi$, and $\Prob^*=\Prob^{\pi^*}$, $\Prob_{\mathrm{ref}}=\Prob^{\pref}$. The normalized occupancy is
$d^\pi_\rho(z)=(1-\gamma)\sum_{t\ge0}\gamma^t\Prob(z_t=z)$, where the probability is for the chain without killing; it equals
$\Prob^\pi(z_T=z)$ (Lemma~\ref{lem:occ}). We write $d^*=d^{\pi^*}_\rho$ and $\dref=d^{\pref}_\rho$. The suboptimality of any
policy is a KL divergence between path laws:

\begin{lemma}[suboptimality is a path KL]\label{lem:pathkl}
For every policy $\pi$, history-dependent ones included, $\SubOpt_\rho(\pi)=\beta\KL(\Prob^\pi\|\Prob^*)$.
\end{lemma}

This is the path-space form of the soft Bellman equations and of the variational view of control as inference
\citep{levine2018reinforcement,nachum2017bridging,mei2020global,liu2026sample}; \citet{odonoghue2022connection} gives a state-wise version, and
Appendix~\ref{app:killed} proves it in our notation. All lower bounds below use it to turn a loss at one
state into a KL divergence between action distributions.

\paragraph{Data.} We consider two models of offline data.
\begin{itemize}
\item[(D1)] \emph{Fixed design.} The data are $n$ i.i.d.\ tuples $(s_i,a_i,R_i,s'_i)$ with $(s_i,a_i)\sim\mu$ for a distribution $\mu$ on
$\cZ$ that does not depend on $(P,r)$, $s'_i\sim P(\cdot|s_i,a_i)$, and $R_i\in[0,1]$ with mean $r(s_i,a_i)$, independent across tuples
given the pairs.
\item[(D2)] \emph{Reference trajectories.} The data are the first $n$ steps of a stream generated by $\pref$: $s_0\sim\rho$; at step $t$ the
data record $(s_t,a_t,R_t,s'_t)$ with $a_t\sim\pref(\cdot|s_t)$ and, given the past and $(s_t,a_t)$, the pair $(R_t,s'_t)$ drawn from a law
that depends only on $(s_t,a_t)$, with $s'_t\sim P(\cdot|s_t,a_t)$ and $R_t\in[0,1]$ of mean $r(s_t,a_t)$; then, with probability $1-\gamma$, the next state is redrawn from
$\rho$ (a reset, which is recorded) and otherwise $s_{t+1}=s'_t$.
\end{itemize}
Between resets the stream of (D2) is a killed path of $\pref$, so the expected number of steps at a pair $z$ is about $n\dref(z)$
(Lemma~\ref{lem:traj}). Theorem~\ref{thm:plugin} also covers the model (D2$'$) of $n$ i.i.d.\ tuples with $(s_i,a_i)\sim\dref$. We
use (D2) rather than (D2$'$) as the model of reference data because the law of a pair drawn from $\dref$ depends on $P$, so the
labels of (D2$'$) carry information about the kernel that no recorded transition carries.

\paragraph{Coverage.} For a data distribution $\mu$ on $\cZ$, the concentrability with respect to the regularized optimum is
$C^*(\mu)=\max_zd^*(z)/\mu(z)$, with $0/0=0$; for reference data we write $C^*(\pref)=C^*(\dref)$. Since $\pi^*$ puts positive mass
on every action, $C^*(\mu)<\infty$ requires $\mu>0$ at every pair of every state that $\pi^*$ reaches. The upper bounds therefore
use a set form: for $E\subseteq\cZ$, $C_\Sigma(E)=\sum_{z\in E}d^*(z)/\mu(z)$ and the uncovered mass $d^*(E^c)$. Always
$C_\Sigma(\cZ)\le KC^*(\mu)$. For contextual bandits $C_\Sigma$ is the coverage coefficient of \citet{zhao2026towards}.

\paragraph{Estimators.} The plug-in estimator $\pi_{\mathrm{PI}}$ is the soft-optimal policy of the empirical model
$(\hat P,\hat r)$, with $\hat Q=0$ at pairs without samples. Pessimistic soft value iteration (PSVI) is defined in
Section~\ref{sec:upper}.

\section{Upper bounds}\label{sec:upper}

\paragraph{Pessimistic soft value iteration.} Let $n_z$ be the number of samples at the pair $z$, and $\hat r(z)$, $\hat P(\cdot|z)$ the empirical
mean reward and next-state law at seen pairs ($n_z\ge1$). For $V:\cS\to[0,H]$, let $\hat\sigma_z(V)$ be the empirical standard deviation (normalized by $n_z$) of the
values $R_i+\gamma V(s'_i)$ observed at $z$. Pessimistic soft value iteration (PSVI) with a penalty $b$ returns $\hat\pi\propto\pref e^{\hat Q/\beta}$,
where $\hat Q$ is the fixed point of
\[
\mathcal T(Q)(z)=\min\big\{H,\ \max\{0,\ \hat r(z)+\gamma\hat PV_Q(z)-b_z(V_Q)\}\big\}\ \ (n_z\ge1),\quad \mathcal T(Q)(z)=0\ \ (n_z=0),
\]
with $V_Q(s)=\beta\log\sum_a\pref(a|s)e^{Q(s,a)/\beta}$. We use two penalties, with $\iota=\log(8KH\bar n^2/\delta)$ and $\bar n$ a bound on the counts ($\bar n=n$ under (D1)):
\begin{align*}
b^{\mathrm H}_z&=\sqrt2H\Big(\tfrac12\sqrt{\iota/n_z}+\iota/n_z\Big)&&\text{(Hoeffding)},\\
b^{\mathrm B}_z(V)&=\max\Big\{4\hat\sigma_z(V)\sqrt{\iota/n_z},\ 16H\iota/n_z\Big\}&&\text{(Bernstein)}.
\end{align*}

\begin{lemma}\label{lem:contraction}
With either penalty, $\mathcal T$ maps $\mathbb R^{\cZ}$ into $[0,H]^{\cZ}$ and is monotone and a $\gamma$-contraction in sup norm on $[0,H]^{\cZ}$. Its
fixed point is unique, and value iteration converges to it from any starting point.
\end{lemma}

Under the Bernstein penalty, the floor $16H\iota/n_z$ makes the map contract: for $V\in[0,H]^{\cS}$, where the variance term is active, the derivative
of $\hat r+\gamma\hat PV-b^{\mathrm B}(V)$ with respect to $V(s')$ is a nonnegative weight, and the weights sum to $\gamma$. With the additive
penalty $\sqrt{\iota\hat\sigma^2/n_z}+H\iota/n_z$ the weights can be negative and the map need not contract (Remark~\ref{rem:additive}). A floor of this kind is used for unregularized pessimistic value iteration with independent samples
by \citet{li2024settling}, who also prove contraction on the box.

\paragraph{PSVI solves a perturbed problem.} Let $\zeta=r+\gamma P\hat V-\hat Q$, with $\hat V=V_{\hat Q}$. By construction $\hat Q=(r-\zeta)+\gamma P\hat V$, so $\hat\pi$ is
the soft-optimal policy for the reward $r-\zeta$ and the true kernel. On the event
\[
\mathcal G=\big\{|\hat r(z)-r(z)+\gamma(\hat P-P)\hat V(z)|\le b_z(\hat V)\ \text{for every seen }z\big\},
\]
$0\le\zeta\le\min\{H,2b\}$ at seen pairs and $0\le\zeta\le H$ at unseen ones (Proposition~\ref{prop:perturbed}). The analysis of the slack rests on
the following lemma. For a slack $\zeta:\cZ\to[0,\infty)$, let $\hat\pi$ be the soft-optimal policy for the reward $r-\zeta$, with the true kernel,
reference policy and temperature, and let $Z=\sum_{t\le T}\zeta(z_t)$ be the slack accumulated along the killed path.

\begin{lemma}[path-space pessimism]\label{lem:pathpess}
With $\hat\Prob=\Prob^{\hat\pi}$, $\E_*$ the expectation under $\Prob^*$, and $\varphi(x)=x-1+e^{-x}$,
\[
\SubOpt_\rho(\hat\pi)\ \le\ \E_*Z-\hat\E Z\ \le\ \beta\,\E_*\varphi(2Z/\beta)\ \le\ 2\,\E_*\min\{Z^2/\beta,\,Z\}.
\]
\end{lemma}

The lemma holds for every kernel and every slack $\zeta\ge0$, and the expectation is under the path law of $\pi^*$, which
single-policy coverage controls. In a one-step problem the quadratic bound holds with constant $1$ instead of our $2$
\citep{zhao2026towards,ji2026optimal}; the middle bound of the lemma is within a factor $2$ of the best possible (Remark~\ref{rem:pathpess}). In online episodic KL-regularized RL, \citet{zhao2025logarithmic} analyze optimism and bound the
suboptimality by a sum over steps of squared future Bellman errors under the learned policy. Lemma~\ref{lem:pathpess} instead has a single
squared sum along the path, taken under the law of $\pi^*$, which coverage of $\pi^*$ controls. The proof (Appendix~\ref{app:killed}) combines Lemma~\ref{lem:pathkl} with the
Donsker--Varadhan inequality.

\paragraph{Rates.} For reference trajectories (D2), PSVI is run on a subsample (Appendix~\ref{app:traj}): of the first $2M$ complete
segments, $M=\lfloor n/(4H)\rfloor$, the even ones set trimmed counts $\tilde n_z$ (lower confidence bounds on the counts in the odd ones), and PSVI uses the first $\tilde n_z$ odd-segment samples at each pair,
with $\bar n=\lceil2MH\rceil$ in $\iota$. We call this the subsampled estimator.

\begin{theorem}[$H^4$ rate for every temperature]\label{thm:h4}
(a) Under (D1), with probability at least $1-2\delta$, PSVI with $b^{\mathrm H}$ satisfies, for every set $E\subseteq\{z:n\mu(z)\ge8\iota\}$,
\[
\SubOpt_\rho(\hat\pi)\ \le\ \cu \min\Big\{\sqrt{\frac{H^4C_\Sigma(E)\,\iota}n}+\frac{H^2C_\Sigma(E)\,\iota}n,\ \ \frac{H^4C_\Sigma(E)\,\iota\,\Lambda}{\beta n}\Big\}+4H^2d^*(E^c),
\]
where $\Lambda=2+\log^+(n/(8\iota C_\Sigma(E)))$, the product $C_\Sigma(E)\Lambda$ is read as $0$ when $C_\Sigma(E)=0$, and $\cu$ is an absolute constant.
(b) Under (D2) with $n\ge68H$, with probability at least $1-2\delta-e^{-M/2}$, the subsampled estimator with $b^{\mathrm H}$ satisfies the same bound with
$\mu=\dref$ and $n$ replaced by $n'=MH/4\ge n/17$, for every $E\subseteq\{z:n\dref(z)\ge17408H\iota\}$.
\end{theorem}

\begin{theorem}[$H^3$ rate for $\beta\ge1.4H$]\label{thm:h3}
Let $\beta\ge1.4H$ and $L_n=2+\log n$. (a) Under (D1), with probability at least $1-2\delta$, PSVI with $b^{\mathrm B}$ satisfies,
for every set $E\subseteq\{z:n\mu(z)\ge16\iota\}$,
\[
\SubOpt_\rho(\hat\pi)\ \le\ \cu \Big[\frac{H^3C_\Sigma(E)\,\iota\,L_n}{\beta n}+\frac{H^3}\beta\,d^*(E^c)\Big];
\]
in particular $\SubOpt_\rho(\hat\pi)\le \cu'H^3C_\Sigma\,\iota\,L_n/(\beta n)$ with $C_\Sigma=\sum_zd^*(z)/\mu(z)$. (b) Under (D2) with $n\ge68H$, with probability at
least $1-2\delta-e^{-M/2}$, the subsampled estimator with $b^{\mathrm B}$ satisfies the first bound of (a) with $\mu=\dref$, $n'$ in place of $n$ (also in $L_n$) and a larger constant, for every
$E\subseteq\{z:n\dref(z)\ge17408H\iota\}$.
\end{theorem}

\begin{corollary}\label{cor:h3}
Let $\beta\ge1.4H$. There is an absolute constant $c$ such that, under (D1), $n\ge c\,KC^*(\mu)H^3\iota L_n/(\beta\epsilon)$ samples give
$\SubOpt_\rho(\hat\pi)\le\epsilon$ with probability at least $1-2\delta$.
\end{corollary}

With $\delta$ of order $\epsilon/H^2$ the corollary also bounds the expected loss (Appendix~\ref{app:cor}), and Theorem~\ref{thm:lb-design} shows that
$\Omega(KCH^3/(\beta\epsilon))$ samples are necessary for expected loss $\epsilon\lesssim\gamma^2H^2/\beta$, so the rate is minimax up to $\iota L_n$. For every $\beta$, the
same rate holds after a burn-in: if $n\mu(z)\ge2c_{\mathrm{burn}}H\iota L_n$ on $E$ for an absolute
$c_{\mathrm{burn}}$, the bound of Theorem~\ref{thm:h3}(a) holds with uncovered term $(H^2+H^3/\beta)d^*(E^c)$ (Proposition~\ref{prop:burnin} in Appendix~\ref{app:burnin}); it matches the lower bound when $\epsilon\lesssim\beta$.

\paragraph{Proof ideas.} On $\mathcal G$, Lemma~\ref{lem:pathpess} bounds the loss by $2\E_*\min\{Z^2/\beta,Z\}$. With the Hoeffding penalty
the slack at a covered pair is of order $H\sqrt{\iota/n_z}$; along a path of expected length $H$ it accumulates to order $H^2\sqrt{\iota/n}$, and the
square of that gives $H^4$. The event $\mathcal G$ follows from a leave-one-out argument under (D1) and from independent tapes of
transitions at each pair under (D2). In Theorem~\ref{thm:h3}, the Bernstein penalty replaces the
factor $H$ in the slack by the standard deviation of the next value of $\hat V$. An energy argument bounds the path sums of that variance and of the action
variance $\Gamma$ of the error $Q^*-\hat Q$ by $O(H^2)$ when $\beta\ge1.4H$; the range part of the penalty enters only linearly, so no count burn-in is needed. Lemma~\ref{lem:pathpess} gives the rate
directly up to $\beta=3.6H$; for larger $\beta$ we use a Donsker--Varadhan step, which bounds the loss by $O(\E_*\Gamma/\beta)$. Full proofs are in Appendix~\ref{app:upper}.

\section{A lower bound for fixed designs}\label{sec:lb-design}

\begin{theorem}\label{thm:lb-design}
Let $\gamma\ge4/7$, $\beta>0$, $C\ge4$, $J\ge1$ and $K=4J+1$, and let
$\epsilon<\frac12\min\{0.00263\,\gamma^2H^2/\beta,\ 0.0281\,\gamma H\}$. There is a family of MDPs with $K$ pairs, one initial
distribution $\rho$ and one design $\mu$, the same for every member, with $d^*\le C\mu$ for every member (so $C^*(\mu)\le C$ and
$C_\Sigma(\cZ)\le KC$), such that every estimator that uses $n$ samples of (D1) and has $\E[\SubOpt_\rho]\le\epsilon$ on every member
needs
\[
n\ \ge\ 4.3\cdot10^{-5}\,\gamma^3\,\frac{JCH^3}{\epsilon\max\{\epsilon,\beta\}}\ \ge\ 1.6\cdot10^{-6}\,\frac{KCH^3}{\epsilon\max\{\epsilon,\beta\}} .
\]
\end{theorem}

The bound is the generative-model lower bound of \citet{liu2026sample} with the design's concentrability as a factor. Its
dependence on $K$, $C$, $\epsilon$ and $\beta$ is that of the minimax rate for offline KL-regularized bandits \citep{ji2026optimal},
and the extra factor $H^3$ comes from the horizon. 

\paragraph{Construction and proof idea.} The family repeats $J$ two-armed gadgets \citep{liu2026sample} (Figure~\ref{fig:constructions} in Appendix~\ref{app:lower}). A gadget has a decision
state $x$ with two actions and uniform reference policy; action $i$ leads to a loop state $y_i$ with reward $1$, which it leaves
for an absorbing zero-reward state with probability $1-p_i$, where $p_{1,2}=p_0\pm\vartheta\Delta$ for a sign $\vartheta$ and a constant
$p_0$. The regularized optimum at $x$ is $\pi^*(1|x)=\sigma(\vartheta u)$ with $\sigma$ the logistic function and
$\beta u\ge\frac98\gamma^2H^2\Delta$, and every path from $\rho$ visits one decision state, at time $0$. By Lemma~\ref{lem:pathkl} the
loss of any policy is the average over gadgets of $\beta\KL(\pi(\cdot|x_j)\|\sigma(\theta_ju))$. Because the design is common to the
family, the KL divergence between the data laws of two members is $n\sum_z\mu(z)\KL(P_\theta(\cdot|z)\|P_{\theta'}(\cdot|z))$, which
only the loop pairs carry. Assouad's argument over the signs, with $\Delta$ chosen so that the information per gadget is a
constant, gives the bound. The design puts mass of order $1/(JC)$ on each loop pair, enough for $d^*\le C\mu$. The full proof is in Appendix~\ref{app:lower}.

\section{Coverage of the regularized optimum by the reference policy}\label{sec:coverage}

\begin{theorem}\label{thm:coverage}
For every MDP, initial distribution $\rho$, reference policy, temperature $\beta$, and every event $E$ of the killed path,
\[
\Prob_{\mathrm{ref}}(E)\ \ge\ e^{-c_1H/\beta}\,\Prob^*(E)^{1+c_2H/\beta},\qquad c_2=6,\ \ c_1=2.75 .
\]
In particular $\dref(E)\ge e^{-c_1H/\beta}d^*(E)^{1+c_2H/\beta}$ for every set $E\subseteq\cZ$. More generally, the bound holds for every
$c_2>\sqrt{32(e-2)}\approx4.79$ with $c_1=c_1(c_2)$ given in Appendix~\ref{app:coverage}; for example $(c_2,c_1)=(5,4.43)$ or $(10,1.77)$.
\end{theorem}

\begin{proof}[Proof outline]
The likelihood ratio of the two path laws is $L=d\Prob^*/d\Prob_{\mathrm{ref}}=\prod_{t\le T}\pi^*(a_t|s_t)/\pref(a_t|s_t)$. For
$\alpha>0$, H\"older's inequality gives
\[
\Prob^*(E)=\E_{\mathrm{ref}}[L\mathbf 1_E]\ \le\ (\E_*L^\alpha)^{1/(1+\alpha)}\,\Prob_{\mathrm{ref}}(E)^{\alpha/(1+\alpha)},
\]
so it suffices to bound $\E_*L^\alpha$ at $\alpha=\beta/(6H)$, that is, a R\'enyi divergence between the path laws. Along the path,
$\beta\log(\pi^*/\pref)(a_t|s_t)=Q^*(z_t)-V^*(s_t)$, and writing $\gamma PV^*(z_t)=V^*(s_{t+1})+\xi_{t+1}$ the values telescope:
$\beta\log L=\sum_{t\le T}r(z_t)+\Xi-V^*(s_0)$, where $\Xi=\sum_t\xi_{t+1}$ is a martingale with increments bounded by $H$. The reward sum is
at most the path length, whose exponential moments are those of a geometric variable. For $\Xi$ an exponential supermartingale
reduces the problem to the exponential moment of the path's total conditional variance $\sum_t\E[\xi_{t+1}^2|z_t]$, which Khasminskii's argument
controls because its conditional expectation from every pair is at most $2H^2$ (Lemma~\ref{lem:tvbound}, a soft version of the
total-variance argument of \citet{azar2013minimax}). This gives $\log\E_*L^{\beta/(6H)}\le0.4571$ (Proposition~\ref{prop:mgf}), and the theorem follows. Details are in
Appendix~\ref{app:coverage}.
\end{proof}

We can turn Theorem~\ref{thm:coverage} into a bound on the coverage coefficient of reference data by taking $E=\{z_T\in F\}$ for
the top-$j$ pairs $F$ in the ratio $d^*/\dref$. Let $S_r$ be the number of states reachable from $\rho$, $E_\tau=\{z:d^*(z)>\tau\}$ and
$a=c_2H/\beta$.

\begin{corollary}\label{cor:topj}
With $\mu=\dref$,
\[
C_\Sigma(E_\tau)\ \le\ \min\Big\{e^{c_1H/\beta}\tau^{-a}\sum_{j=1}^{|E_\tau|}j^{-a},\ \ |E_\tau|\,e^{(2S_r-1)H/\beta}\Big\}.
\]
For $a<1$ the first term is at most $e^{c_1H/\beta}\tau^{-a}|E_\tau|^{1-a}/(1-a)$.
\end{corollary}

The second term depends only on the number of reachable states.

\begin{proposition}[finite-state cap]\label{prop:cap}
$C^*(\pref)=\max_zd^*(z)/\dref(z)\le e^{(2S_r-1)H/\beta}$.
\end{proposition}

The proof (Appendix~\ref{app:coverage}) uses the Markov chain tree theorem. Hence the ratio
$d^*(E)/\dref(E)$ can grow without bound as $d^*(E)\to0$ only if the number of reachable states grows. With deterministic transitions and
a fixed horizon the bound holds without the power.

\begin{proposition}[deterministic transitions]\label{prop:det}
Consider the finite-horizon version of the problem: $H$ stages, no discounting, rewards in $[0,1]$, and the regularized optimum
$\pi^*_h\propto\pref e^{Q^*_h/\beta}$ with the stage-dependent soft values $Q^*_h=r+PV^*_{h+1}$, $V^*_{H+1}=0$. With deterministic transitions,
$\Prob^*(\tau)\le e^{H/\beta}\Prob_{\mathrm{ref}}(\tau)$ for every trajectory $\tau$.
\end{proposition}

With deterministic transitions the martingale $\Xi$ vanishes, while stochastic transitions or a random effective horizon make it nonzero.
Even with deterministic transitions, discounting alone (a random horizon) makes the likelihood ratio unbounded. Take one state with two
self-loop actions, rewards $1$ and $0$, and a uniform reference: a killed path that takes the rewarded action $N$ times has likelihood ratio $(2p)^N$,
with $p=1/(1+e^{-1/\beta})>\frac12$, which is unbounded in $N$. With stochastic transitions a power can be necessary, at least for $\beta\ge H$:

\begin{lemma}[chains]\label{lem:chain}
Consider a chain of states $c_0,\dots,c_{m-1}$ with zero rewards, started at $c_0$. At each $c_k$ the action go has reference
probability $\frac12$ and moves to $c_{k+1}$ with probability $\frac12$ (from $c_{m-1}$, to an end state with any value in $[0,H]$) and
otherwise to an absorbing state of value $H$; the action quit moves to an absorbing zero-reward state. If $\beta\ge H$, the
probabilities $\delta_*$ and $\delta_{\mathrm{ref}}$ that the killed chains of $\pi^*$ and $\pref$ reach the end state satisfy
$(\gamma/4)^m\le\delta_*\le0.42^m$ and $\delta_{\mathrm{ref}}\le\delta_*^{1+\kappa H/\beta}$, where $\kappa=3\gamma/(16\log(4/\gamma))\ge0.045$.
\end{lemma}

At each chain state $\pi^*$ chooses go with probability larger than the reference probability $\frac12$ by a factor
$e^{\Omega(H/\beta)}$, while the reach probability shrinks by a constant factor per state (Appendix~\ref{app:reference}). For $\beta\ge H$ the exponent
of Theorem~\ref{thm:coverage} is therefore right up to a constant factor.

\section{The price of reference data}\label{sec:price}

We now show that this loss of coverage costs samples. Both theorems below use reference trajectories (D2) and count
the data in transitions.

\begin{theorem}[power-law price, $\beta\ge H$]\label{thm:power}
Let $\gamma\ge4/7$, $\beta\ge H$, $\kappa=3\gamma/(16\log(4/\gamma))$, and let $\epsilon>0$ with
$\delta_\epsilon:=1518\,\epsilon\beta/(\gamma^3H^2)<\gamma/4$. Let $\bar m=1+\log(1/\delta_\epsilon)/\log(1/0.42)$. For every
$K\ge2(2\bar m+4)$ there is a family of MDPs with $K$ pairs and one initial state such that every estimator that uses $n$ transitions
of (D2) and has $\E[\SubOpt_\rho]\le\epsilon$ on every member needs
\[
n\ \ge\ c\,\frac{KH^3}{\epsilon\beta}\Big(\frac{c'H^2}{\epsilon\beta}\Big)^{\kappa H/\beta}-H,\qquad c\ge1.6\cdot10^{-6},\ \ c'\ge1.7\cdot10^{-5}.
\]
\end{theorem}

The construction puts one chain of Lemma~\ref{lem:chain} in front of $\Theta(K)$ gadgets (Figure~\ref{fig:constructions}). $\pi^*$ reaches the gadgets with probability
$\delta_*\approx\epsilon\beta/H^2$, so they contribute about $\epsilon$ to the loss, and the reference data with probability at most $\delta_*^{1+\kappa H/\beta}$. The values
on the chain are symmetric in the gadget signs, so a learner either reaches the gadgets often enough to pay the two-point loss there or
loses a constant fraction of $\beta\delta_*$ by deviating on the chain.

For $\beta<H$ a gate in front of each gadget gives coverage $C$ up to an exponential in $H/\beta$:

\begin{theorem}[gateway family]\label{thm:gate}
Let $\gamma\ge4/7$, $\beta>0$ and $C\ge4$ with $C/2-1\le\exp(\gamma^2H/(3\beta))$. Let $J\ge1$, $K=6J+2$ and
$\epsilon<\frac\gamma8\min\{0.00263\,\gamma^2H^2/\beta,\ 0.0281\,\gamma H\}$. There is a family of MDPs with $K$ pairs, one initial
distribution and one reference policy, with $C/4\le C^*(\pref)\le C$ for every member, such that every estimator that uses $n$
transitions of (D2) and has $\E[\SubOpt_\rho]\le\epsilon$ on every member needs
\begin{gather*}
n+H\ \ge\ JC\min\Big\{7.2\cdot10^{-5}\,\frac{\gamma^2H^3}{\epsilon\beta},\ 2.7\cdot10^{-6}\,\frac{\gamma^3H^3}{\epsilon^2}\Big\},\\
\text{so}\quad n\ \ge\ 6.3\cdot10^{-8}\,\frac{KCH^3}{\epsilon\max\{\epsilon,\beta\}}-H .
\end{gather*}
\end{theorem}

The gate enters its gadget with reference probability $2/C$ and the optimum enters with probability at least $\frac12$, so
$C/4\le C^*(\pref)\le C$; skipping a gate loses a constant fraction of $\gamma^2H$. The fixed-design lower bound of Theorem~\ref{thm:lb-design} thus
also holds for reference data with the same coverage. Proofs of both theorems are in Appendix~\ref{app:reference}.

Conversely, by Theorem~\ref{thm:coverage} the pairs with $d^*(z)>\tau$ meet the count condition of Theorems~\ref{thm:h4}(b) and~\ref{thm:h3}(b) once
$\tau$ is of order $(H\iota/n)^{1/(1+c_2H/\beta)}$, and the other pairs carry mass at most $K\tau$, so the following corollary needs no coverage assumption.

\begin{corollary}[reference data without a coverage assumption]\label{cor:refdata}
Let $c_1=2.75$ and $c_2=6$ be the constants of Theorem~\ref{thm:coverage}, $a=c_2H/\beta$ and $\tau_n=(17408H\iota e^{c_1H/\beta}/n)^{1/(1+a)}$. Under (D2) with $n\ge68H$, with probability
at least $1-2\delta-e^{-M/2}$, the subsampled estimator with $b^{\mathrm H}$ satisfies $\SubOpt_\rho(\hat\pi)\le \cu K\tau_n(H^2+H^3L_n/\beta)$ for every $\beta$,
and with $b^{\mathrm B}$ it satisfies $\SubOpt_\rho(\hat\pi)\le \cu K\tau_n(H^3+H^2L_n)/\beta$ for $\beta\ge1.4H$. In particular, for $\beta\ge1.4H$,
with $b^{\mathrm B}$, $n\ge c\,H\iota\,(KH^3L_n/(\beta\epsilon))^{1+c_2H/\beta}$ transitions give $\SubOpt_\rho(\hat\pi)\le\epsilon$ with that probability. Here
$\cu\ge4$ and $c$ are absolute constants.
\end{corollary}

The proof is in Appendix~\ref{app:refdata}. The exponent of $1/\epsilon$ is $1+O(H/\beta)$. For $\beta\ge H$, Theorem~\ref{thm:power} shows that
bounds polynomial in $K$ need the exponent $1+\Omega(H/\beta)$ (its family has $K$ of order $\log(1/\epsilon)$, and it bounds the expected loss).

\section{Pessimism and reference data}\label{sec:plugin}

With a generative model the plug-in estimator matches the lower bound \citep{liu2026sample}, but with reference data it can fail. The
instance $I(m,\gamma)$ has a decision state $x$ and absorbing states $G$ (reward $1$) and $B$ (reward $0$), $\rho$ the point mass at $x$, and reward
$0$ at $x$. At $x$ a main action with $\pref=\frac12$ moves to $G$ with probability $\frac34$, and each of $m$ rare actions, with
$\pref=\frac1{2m}$, moves to $G$ with probability $\frac12$; otherwise the action moves to $B$.

\begin{theorem}[the plug-in fails with reference data]\label{thm:plugin}
Let $\gamma\ge\frac12$, $n\ge2000H$ and $m=\lceil n/H\rceil$; let $k_0$ be the least integer $k\ge8$ with $m(e/(k+1))^{k+1}\le0.05$, and let
\[
\beta\ \le\ \min\Big\{\frac{\gamma H}{k_0\log80},\ \frac{\gamma H}{4\log m}\Big\}.
\]
On $I(m,\gamma)$, which has $K=\lceil n/H\rceil+3$ pairs and $C_\Sigma(\cZ)\le6$ for $\mu=\dref$, the plug-in estimator has
$\E[\SubOpt_\rho(\pi_{\mathrm{PI}})]\ge0.09\,\gamma H$ under (D2) and under (D2$'$).
\end{theorem}

Here $k_0=O(\log m/\log\log m)$, so the condition is $\beta\lesssim\gamma H/\log n$, and it implies $\beta\le\gamma H/35$; for $\gamma H/\beta\ge40$ the
first condition can be dropped (Remark~\ref{rem:plugin}). Each rare action is sampled about once; a constant fraction is sampled exactly once
with a move to $G$, the plug-in values them at $\gamma H$ against at most $\frac78\gamma H$ for the main action, and its soft-max puts at least half
its mass on them, where $\pi^*$ puts at most $e^{-\gamma H/(4\beta)}$. The proof (Appendix~\ref{app:plugin}) conditions on the number of transitions
from $x$, given which the two data models agree. This is the KL-regularized counterpart of the failure of the empirical best arm in
bandits even when $C^*$ is close to $1$ \citep[Proposition~1]{rashidinejad2021bridging}.

PSVI, on the other hand, learns on these instances. Under the condition of Theorem~\ref{thm:plugin}, the main action and the two absorbing
states have bounded coverage and miss at most $1/n$ of the optimum's occupancy. Theorem~\ref{thm:h4} therefore bounds the loss of PSVI under
(D2$'$), which on a fixed instance is (D1) with $\mu=\dref$, and of the subsampled estimator under (D2), by quantities that tend to $0$ as $n$
grows at fixed $H$, for every $\beta$ that Theorem~\ref{thm:plugin} allows (Appendix~\ref{app:plugin} gives the bounds).

\section{In simulation, the variance-aware penalty saves a power of \texorpdfstring{$H$}{H}}\label{sec:experiments}

\begin{figure}[ht]
\centering
\includegraphics[width=0.92\linewidth]{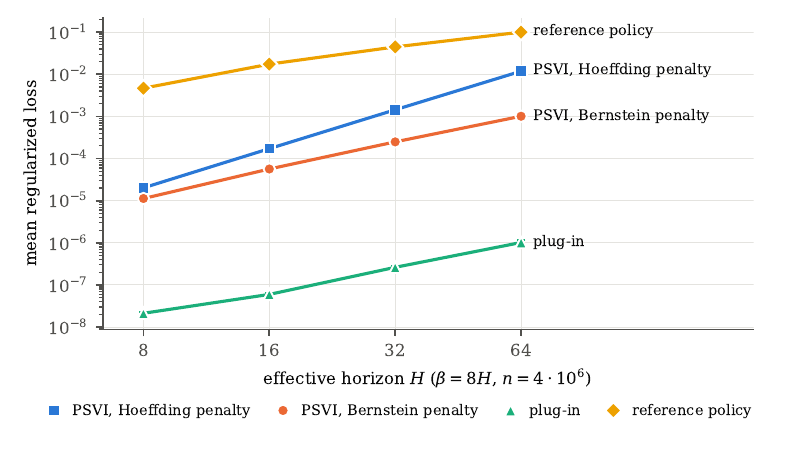}
\caption{The loop family ($\beta=8H$, $n=4\cdot10^6$): mean regularized loss of PSVI with the Hoeffding and the floored Bernstein
penalties and of the plug-in estimator, over 120 datasets per $H$ (standard errors of the pooled means at most \CNineteenPooledSEPessimistic\% of the mean for PSVI and
\CNineteenPooledSEPlugin\% for the plug-in), and the exact loss of the reference policy. The least-squares slopes of the pooled means over the four
horizons are $\CNineteenPooledSlopeHoe$ for the Hoeffding penalty and $\CNineteenPooledSlopeBern$ for the Bernstein penalty; on this family at fixed $n$ the fast branch
of Theorem~\ref{thm:h4} grows as $H^3$ and the bound of Theorem~\ref{thm:h3} as $H^2$.}
\label{fig:c19c}
\end{figure}

To see whether the two rates differ at moderate sample sizes, we simulated PSVI on a family where a well-sampled loop lies on the optimum's
path (Appendix~\ref{app:checks}), at $\beta=8H$ and a fixed $n$. On this family the $H^4/(\beta n)$ bound of Theorem~\ref{thm:h4} grows as $H^3$ and the
bound of Theorem~\ref{thm:h3} as $H^2$. The measured losses grow at nearly the same rates, with fitted exponents of $\CNineteenPooledSlopeHoe$ for the
Hoeffding penalty and $\CNineteenPooledSlopeBern$ for the Bernstein penalty (Figure~\ref{fig:c19c}), and the local slope of the latter falls toward $2$ as $H$
grows. The plug-in estimator, which has no penalty, loses less than both here, so this family says nothing about the need for pessimism.
Appendix~\ref{app:checks} also reports exact computations for the constructions of the lower bounds and numerical evaluations of the main
inequalities on small random instances.

\section{Discussion}\label{sec:discussion}

\subsection{Limitations}\label{sec:limitations}

The main limitation is the temperature range of the $H^3$ rate. Without a burn-in we prove it for $\beta\ge1.4H$, where $\pi^*$ stays within a factor
$e^{H/\beta}\le2.043$ of $\pref$ at every state; the proof extends down to about $1.31H$, with a constant that grows without bound
(Remark~\ref{rem:h3range}). For smaller temperatures we have the $H^4$ rate, or $H^3$ after a burn-in with a large constant and only when
$\epsilon\lesssim\beta$ (Proposition~\ref{prop:burnin}). At small temperatures the problem approaches unregularized RL, for which \citet{li2024settling}
obtain the $H^3$ rate with the same kind of penalty. Closing the gap would need a bound on the variance of the error along the optimum's path
that does not go through our energy argument, which fails below about $1.31H$.

Our constants are also large enough that, at the sample sizes we simulate, the bounds exceed the loss of simply returning $\pref$
(Appendix~\ref{app:constants}). Smaller constants would need sharper forms of the inequalities of Appendix~\ref{app:upper}, e.g., a joint control of the
local variance and the energy of the error.

For reference data our guarantees are loose. The subsampled estimator uses about a quarter of the data and treats pairs with fewer than
order $H\iota$ expected visits as uncovered, which can cost a factor $H$; an analysis of PSVI on whole trajectories might avoid both. The exponent
of $1/\epsilon$ is $1+6H/\beta$ in Corollary~\ref{cor:refdata} but $1+\kappa H/\beta$, with a small $\kappa$, in Theorem~\ref{thm:power}, whose extra factor becomes large
only at very high accuracy ($H/\epsilon$ above about $10^7$ with our constants). Matching the exponents needs a sharper coverage bound, a harder
family of instances, or both.

\subsection{Conclusion}\label{sec:conclusion}

Pessimistic soft value iteration learns from a fixed design at the minimax rate, up to logarithmic factors, when $\beta\ge1.4H$. With stochastic
transitions the likelihood ratio between $\pi^*$ and $\pref$ can be unbounded, yet $\pref$ still covers the optimum's path events up to a power law,
and in that temperature range this is enough to learn from reference data with a sample size polynomial in $1/\epsilon$ (Corollary~\ref{cor:refdata}).
For $\beta\ge H$ a polynomial increase over the fixed-design rate is sometimes necessary.
We think analyses of learning from reference data---fine-tuning included---are better built on path-level bounds of this kind than on
per-step likelihood ratios, which compound with the horizon. We do not know whether pessimism is needed at large temperatures. Natural next
steps are the $H^3$ rate below about $1.31H$ without a burn-in and matching exponents for reference data.

\bibliographystyle{plainnat}
\bibliography{refs}

\appendix
\section{Killed chains, the path-KL identity, and path-space pessimism}\label{app:killed}

For a policy $\pi$ and an initial distribution $\rho$, the killed chain draws $s_0\sim\rho$ and, for $t=0,1,\dots$, draws
$a_t\sim\pi(\cdot|s_t)$; then, with probability $1-\gamma$ and independently of everything else, it stops and sets $T=t$;
otherwise it draws $s_{t+1}\sim P(\cdot|s_t,a_t)$. Write $z_t=(s_t,a_t)$ and $N=T+1$. Let $\mathcal F_0=\sigma(s_0)$, let $\mathcal G_t$
be the $\sigma$-field of $(s_0,a_0,\dots,s_t,a_t)$ and of the stopping coins up to time $t-1$, and let
$\mathcal F_{t+1}=\mathcal G_t\vee\sigma(\text{coin }t,\,s_{t+1})$, with the convention that $s_{t+1}=\dagg$, a cemetery state with
value $0$, if the chain stopped at $t$.

\begin{lemma}[occupancy as a killed marginal]\label{lem:occ}
$\Prob^\pi(T=t,\,z_t=z)=(1-\gamma)\gamma^t\,\Prob^\pi_{\mathrm{unkilled}}(z_t=z)$. Hence
$d^\pi_\rho(z)=(1-\gamma)\sum_t\gamma^t\Prob^\pi_{\mathrm{unkilled}}(z_t=z)=\Prob^\pi(z_T=z)$.
\end{lemma}
\begin{proof}
The coins are independent of the actions and transitions, so the chain survives the first $t$ coins with probability
$\gamma^t$, independently of $(z_0,\dots,z_t)$, and stops at the next one with probability $1-\gamma$.
\end{proof}

\begin{lemma}[likelihood ratio]\label{lem:lr}
On the space of finite paths with their coins, $d\Prob^*/d\Prob_{\mathrm{ref}}=L$ with
$L=\prod_{t=0}^T\pi^*(a_t|s_t)/\pref(a_t|s_t)$. In particular $d^*(E)=\E_{\mathrm{ref}}[L\,\mathbf 1\{z_T\in E\}]$ for every
$E\subseteq\cZ$.
\end{lemma}
\begin{proof}
The two path probabilities share the initial law, the transition factors and the coin factors, and differ only in the
action factors. The second statement is Lemma~\ref{lem:occ} for $\pi^*$.
\end{proof}

\begin{lemma}[telescoping]\label{lem:tele}
Let $\xi_{t+1}=\gamma PV^*(z_t)-V^*(s_{t+1})$ for $t\le T$, with $s_{T+1}=\dagg$. Then, pathwise,
\[
\beta\log L=R+\Xi-V^*(s_0),\qquad R=\sum_{t=0}^Tr(z_t)\le N,\qquad \Xi=\sum_{t=0}^T\xi_{t+1}.
\]
Moreover $\E[\xi_{t+1}|\mathcal G_t]=0$, $|\xi_{t+1}|\le H$, and
$v_t:=\E[\xi_{t+1}^2|\mathcal G_t]=\gamma\Var_PV^*(z_t)+\gamma(1-\gamma)(PV^*(z_t))^2\le\Var_PV^*(z_t)+H$.
\end{lemma}
\begin{proof}
$\beta\log(\pi^*/\pref)(a|s)=Q^*(s,a)-V^*(s)=r(s,a)+\gamma PV^*(s,a)-V^*(s)$. Writing $\gamma PV^*(z_t)=\xi_{t+1}+V^*(s_{t+1})$ and
summing over $t\le T$, the terms $V^*(s_{t+1})-V^*(s_t)$ telescope to $V^*(s_{T+1})-V^*(s_0)=-V^*(s_0)$. Given $\mathcal G_t$ on
$\{T\ge t\}$, $s_{t+1}=\dagg$ with probability $1-\gamma$ and otherwise $s_{t+1}\sim P(\cdot|z_t)$, so
$\E[V^*(s_{t+1})|\mathcal G_t]=\gamma PV^*(z_t)$ and $\E[V^*(s_{t+1})^2|\mathcal G_t]=\gamma P(V^{*2})(z_t)$, which gives the variance.
Since $V^*\in[0,H]$ and $\gamma PV^*\in[0,H]$, $|\xi|\le H$; and $\gamma(1-\gamma)(PV^*)^2\le(1-\gamma)H^2=H$.
\end{proof}

\begin{proof}[Proof of Lemma~\ref{lem:pathkl}]
Let $\Prob^\pi$ be the killed path law of $\pi$. As in Lemma~\ref{lem:lr},
$d\Prob^\pi/d\Prob_{\mathrm{ref}}=\prod_{t\le T}\pi(a_t|h_t)/\pref(a_t|s_t)$, where $h_t$ is the history up to $s_t$. By
Lemma~\ref{lem:occ} and the tower property, $V^\pi(\rho)=\E^\pi[R]-\beta\KL(\Prob^\pi\|\Prob_{\mathrm{ref}})$ with
$R=\sum_{t\le T}r(z_t)$. By Lemma~\ref{lem:tele}, $\beta\log(d\Prob^*/d\Prob_{\mathrm{ref}})=R+\Xi-V^*(s_0)$, and $\E^\pi \Xi=0$: the
increments $\xi_{t+1}$ have conditional mean zero given $\mathcal G_t$ under every policy, because the coin and the next state do
not depend on the policy, and $\sum_t\E^\pi[|\xi_{t+1}|\mathbf 1\{T\ge t\}]\le H\,\E N<\infty$ allows exchanging sum and expectation.
Hence $\beta\KL(\Prob^\pi\|\Prob^*)=\beta\KL(\Prob^\pi\|\Prob_{\mathrm{ref}})-\E^\pi R+V^*(\rho)=V^*(\rho)-V^\pi(\rho)$.
\end{proof}

\begin{proof}[Proof of Lemma~\ref{lem:pathpess}]
For a reward $r'$ write $J_{r'}(\pi)=\E^\pi[\sum_{t\le T}r'(z_t)]-\beta\KL(\Prob^\pi\|\Prob_{\mathrm{ref}})$, the regularized value of $\pi$
at $\rho$ for $r'$, so that $J_{r-\zeta}(\pi)=J_r(\pi)-\E^\pi Z$. Lemma~\ref{lem:pathkl} applied to the reward $r-\zeta$ (whose
soft-optimal value is bounded, so the martingale step of its proof applies) gives
$J_{r-\zeta}(\hat\pi)-J_{r-\zeta}(\pi)=\beta\KL(\Prob^\pi\|\hat\Prob)\ge0$ for every policy $\pi$. With $\pi=\pi^*$,
$J_r(\hat\pi)-\hat\E Z\ge J_r(\pi^*)-\E_*Z$, which is the first inequality, since $\SubOpt_\rho(\hat\pi)=J_r(\pi^*)-J_r(\hat\pi)$.

Let $c>0$. The Donsker--Varadhan inequality, applied to $f=c(\E_*Z-Z)$, which is bounded above, gives
$c(\E_*Z-\hat\E Z)\le\log\E_*e^{c(\E_*Z-Z)}+\KL(\hat\Prob\|\Prob^*)$. Since $Z\ge0$ and $\log y\le y-1$,
$\log\E_*e^{c(\E_*Z-Z)}=c\,\E_*Z+\log\E_*e^{-cZ}\le\E_*[cZ-1+e^{-cZ}]=\E_*\varphi(cZ)$. By Lemma~\ref{lem:pathkl},
$\KL(\hat\Prob\|\Prob^*)=\SubOpt_\rho(\hat\pi)/\beta\le(\E_*Z-\hat\E Z)/\beta$. With $c=2/\beta$ this gives
$(\E_*Z-\hat\E Z)/\beta\le\E_*\varphi(2Z/\beta)$, the second inequality. The third follows from $\varphi(x)\le\min\{x^2/2,\,x\}$
for $x\ge0$.
\end{proof}

\begin{remark}\label{rem:pathpess}
Lemma~\ref{lem:pathkl} for $r$ and for $r-\zeta$ also gives the exact relation
$\E_*Z-\hat\E Z=\beta\KL(\hat\Prob\|\Prob^*)+\beta\KL(\Prob^*\|\hat\Prob)$. The constant in Lemma~\ref{lem:pathpess} is within a factor
$2$ of the best possible: in the one-step family $\pi^*=(1-p,p)$, $\zeta=(\beta(\log(1/p)+y),0)$, the ratio
$\SubOpt/(\beta\E_*\varphi(2Z/\beta))$, maximized over $y$, approaches $1/2$ as $p\to0$ (it is $0.493$ at $p=10^{-200}$).
\end{remark}

\section{Proofs for Section~\ref{sec:upper}}\label{app:upper}

Throughout, $\hat Q$ is the fixed point of PSVI's map, $\hat V=V_{\hat Q}$, $\zeta=r+\gamma P\hat V-\hat Q$, $D^*=Hd^*$ is the unnormalized occupancy (so
$\E_*\sum_{t\le T}f(z_t)=\sum_zD^*(z)f(z)$), and $\E_*$ is over the killed chain of $\pi^*$ with the data fixed.

\subsection{The map and the perturbed problem}

\begin{proof}[Proof of Lemma~\ref{lem:contraction}]
The clip puts $\mathcal T(Q)$ in $[0,H]^{\cZ}$ for every $Q\in\mathbb R^{\cZ}$, so every fixed point lies in $[0,H]^{\cZ}$ and value iteration
enters $[0,H]^{\cZ}$ after one step; for $Q\in[0,H]^{\cZ}$, $V_Q\in[0,H]^{\cS}$. The soft value $Q\mapsto V_Q$ and the clip are monotone and
$1$-Lipschitz in sup norm, so it suffices that, for each seen $z$,
$g_z(V)=\hat r(z)+\gamma\hat PV(z)-b_z(V)$ is nondecreasing in each coordinate and $\gamma$-Lipschitz on $[0,H]^{\cS}$. For $b^{\mathrm H}$, which does not
depend on $V$, this is immediate. For $b^{\mathrm B}$, write $n=n_z$, $X_i=R_i+\gamma V(s'_i)\in[0,H]$, $\bar X$ their mean, $\omega=4\sqrt{\iota/n}$ and
$F=16H\iota/n=\omega^2H$, so $g_z(V)=\bar X-\max\{\omega\hat\sigma_z(V),F\}$. On the open set where $\omega\hat\sigma>F$,
\[
\frac{\partial g_z}{\partial V(j)}=\frac\gamma n\sum_{i:\,s'_i=j}\Big(1-\omega\,\frac{X_i-\bar X}{\hat\sigma}\Big),
\]
because $\partial\hat\sigma/\partial V(j)=\frac\gamma n\sum_{i:s'_i=j}(X_i-\bar X)/\hat\sigma$. There $\hat\sigma>\omega H$ and $|X_i-\bar X|\le H$, so each bracket lies in $[0,2]$, and
the brackets average to $1$; the partial derivatives are nonnegative and sum to $\gamma$. On the open set where $\omega\hat\sigma<F$, $g_z=\bar X-F$ has
partial derivatives $\gamma\hat P(j|z)\ge0$, summing to $\gamma$. The function $\hat\sigma_z$ is $1$-Lipschitz in sup norm (the $L^2(\hat P_z)$ norm of the
centered $X$), so $g_z$ is Lipschitz and absolutely continuous along any segment $V+t(V'-V)$. Its derivative there exists for almost
every $t$ and equals the directional derivative from one of the two open sets, or $\gamma\hat P(V'-V)$ on a set where $\omega\hat\sigma=F$ (where
$\frac d{dt}\omega\hat\sigma=0$ almost everywhere). It is nonnegative when $V\le V'$, and at most $\gamma\|V'-V\|_\infty$ in absolute value.
\end{proof}

\begin{remark}\label{rem:additive}
With the additive penalty $\sqrt{\iota\hat\sigma^2/n}+H\iota/n$ the same computation gives brackets $1-\sqrt{\iota/n}(X_i-\bar X)/\hat\sigma$, which are negative
when $\hat\sigma$ is small compared with $H\sqrt{\iota/n}$; the derivatives then need not be nonnegative and their absolute values can sum to
more than $1$, so the map need not be a contraction even on $[0,H]^{\cZ}$. On random empirical laws we found sums up to \CTwentyAdditiveMaxUnclipped\ at points
where the map is not clipped (Appendix~\ref{app:checks}).
\end{remark}

\begin{proposition}[PSVI solves a perturbed problem]\label{prop:perturbed}
$\hat\pi$ is the soft-optimal policy for the reward $r-\zeta$ and the kernel $P$. On the event $\mathcal G$ of Section~\ref{sec:upper},
$0\le\zeta(z)\le\min\{H,2b_z(\hat V)\}$ if $n_z\ge1$ and $0\le\zeta(z)\le H$ if $n_z=0$; moreover $0\le\hat Q\le Q^*$ and $0\le\hat V\le V^*$.
\end{proposition}
\begin{proof}
By definition $\hat Q=(r-\zeta)+\gamma P\hat V$ with $\hat V$ the soft value of $\hat Q$, so $(\hat Q,\hat V)$ is the fixed point of the soft Bellman operator
for $r-\zeta$ and $P$, which is a contraction; hence $\hat\pi$ is soft-optimal for that problem. Since $\hat Q\in[0,H]$, also $\hat V\in[0,H]$ and
$r+\gamma P\hat V\in[0,H]$. If $n_z=0$, $\zeta=r+\gamma P\hat V\in[0,H]$. If $n_z\ge1$ and neither clip is active, $\zeta=(r-\hat r)+\gamma(P-\hat P)\hat V+b\in[0,2b]$
on $\mathcal G$, and $\zeta\le r+\gamma P\hat V\le H$ since $\hat Q\ge0$. If the clip at $0$ is active, $\zeta=r+\gamma P\hat V\ge0$ and
$\zeta<r+\gamma P\hat V-(\hat r+\gamma\hat P\hat V-b)\le2b$. The clip at $H$ is never active on $\mathcal G$, because there $\hat r+\gamma\hat P\hat V-b\le r+\gamma P\hat V\le H$. Finally,
$\zeta\ge0$ gives $\hat Q\le Q^*$ and $\hat V\le V^*$ by monotonicity of the soft Bellman operator in the reward.
\end{proof}

\subsection{The penalty events for a fixed design}\label{app:events}

In this subsection, given counts $(n_z)_z$ with $\max_zn_z\le\bar n$, the samples at each pair are i.i.d.\ tuples $(R,s')$ with $R\in[0,1]$, $\E R=r(z)$
and $s'\sim P(\cdot|z)$ ($R$ and $s'$ may be dependent), independent across pairs; $\iota=\log(8KH\bar n^2/\delta)$. For a seen $z$ and $v\in[0,H]$, let
$Q^{(z,v)}$ be the fixed point of the map that agrees with $\mathcal T$ except that its value at $z$ is the constant $v$.

\begin{lemma}[leave-one-out]\label{lem:loo}
$Q^{(z,v)}$ exists, is unique, does not depend on the samples at $z$, and satisfies $\|Q^{(z,v)}-Q^{(z,v')}\|_\infty\le|v-v'|$; and $\hat Q=Q^{(z,\hat Q(z))}$.
\end{lemma}
\begin{proof}
The modified map is a monotone $\gamma$-contraction by Lemma~\ref{lem:contraction}, and it does not use the samples at $z$. The coordinates other than $z$
of two fixed points $Q,Q'$ for $v,v'$ satisfy $|Q-Q'|\le\gamma\|Q-Q'\|_\infty$, and the coordinate $z$ differs by $|v-v'|$, so $\|Q-Q'\|_\infty\le|v-v'|$. The
original fixed point satisfies the modified equations with $v=\hat Q(z)$, so uniqueness gives the last claim.
\end{proof}

\begin{proposition}[Hoeffding penalty event]\label{prop:eventH}
For PSVI with $b^{\mathrm H}$, conditionally on the counts, $\mathcal G$ holds with probability at least $1-\delta$.
\end{proposition}
\begin{proof}
Fix a seen $z$ and $v$ in a grid $U\subseteq[0,H]$ of spacing $1/n_z^2$, so $|U|\le2H\bar n^2$. Given the counts, $V^{(z,v)}=V_{Q^{(z,v)}}$ is independent of the
samples at $z$, so Hoeffding's inequality for the values $R_i+\gamma V^{(z,v)}(s'_i)\in[0,H]$ gives a deviation of their mean from its expectation of
at most $H\sqrt{\iota/(2n_z)}$, except with probability $2e^{-\iota}$. A union bound over $z$ and $U$ costs $4KH\bar n^2e^{-\iota}\le\delta$. For $v^*=\hat Q(z)$
and $v_g\in U$ nearest to it, Lemma~\ref{lem:loo} gives $\|\hat V-V^{(z,v_g)}\|_\infty\le1/n_z^2$, which changes the deviation by at most $2/n_z^2\le H\iota/n_z$.
With $c_b=\sqrt2$, $H\sqrt{\iota/(2n_z)}=\frac{c_b}2H\sqrt{\iota/n_z}$, so the deviation is at most $b^{\mathrm H}_z$.
\end{proof}

\begin{proposition}[Bernstein penalty event]\label{prop:eventB}
For PSVI with $b^{\mathrm B}$, conditionally on the counts, with probability at least $1-\delta$, for every seen $z$: (i) $|\hat r(z)-r(z)+\gamma(\hat P-P)\hat V(z)|\le b^{\mathrm B}_z(\hat V)$,
so $\mathcal G$ holds; and (ii) $\hat\sigma_z(\hat V)\le\sigma_z(\hat V)+3H\sqrt{\iota/n_z}$, where $\sigma_z(\hat V)$ is the standard deviation of $R+\gamma\hat V(s')$ under the
true law at $z$ with $\hat V$ held fixed.
\end{proposition}
\begin{proof}
Fix a seen $z$ with $n=n_z\ge2$ and $v\in U$, and write $X_i=R_i+\gamma V^{(z,v)}(s'_i)\in[0,H]$, $\sigma^2=\Var X_i$, $s^2=\frac n{n-1}\hat\sigma^2$ for the unbiased
sample variance. Bernstein's inequality for variables bounded above by their mean plus $H$ \citep[Chapter~2]{boucheron2013concentration},
applied to $X_i$ and to $-X_i$, gives $|\bar X-\E X|\le\sigma\sqrt{2\iota/n}+H\iota/(3n)$ except with probability $2e^{-\iota}$. Theorem~10 of
\citet{maurer2009empirical}, applied to $X_i/H$, gives $\sigma\le s+H\sqrt{2\iota/(n-1)}$ and $s\le\sigma+H\sqrt{2\iota/(n-1)}$, each except with probability
$e^{-\iota}$. A union bound over $z$, $U$ and the four inequalities costs $8KH\bar n^2e^{-\iota}=\delta$. For $v^*=\hat Q(z)$ and the nearest $v_g\in U$,
$\|V^{(z,v^*)}-V^{(z,v_g)}\|_\infty\le1/n^2$ changes $\bar X-\E X$ by at most $2/n^2$, and $\hat\sigma$ and $\sigma$ by at most $1/n^2$ each. Using $1/(n-1)\le2/n$,
$s\le\sqrt2\hat\sigma$ and $1/n^2\le H\iota/n$,
\[
|\bar X-\E X|\big|_{v^*}\le\big(\sqrt2(\hat\sigma(v^*)+\tfrac1{n^2})+2H\sqrt{\iota/n}\big)\sqrt{2\iota/n}+\frac{H\iota}{3n}+\frac2{n^2}\le2\hat\sigma(v^*)\sqrt{\iota/n}+7.2\,\frac{H\iota}n,
\]
while $b^{\mathrm B}_z\ge\frac12\cdot4\hat\sigma\sqrt{\iota/n}+\frac12\cdot16H\iota/n$, which gives (i). For (ii),
$\hat\sigma(v^*)\le\hat\sigma(v_g)+\frac1{n^2}\le s(v_g)+\frac1{n^2}\le\sigma(v_g)+2H\sqrt{\iota/n}+\frac1{n^2}\le\sigma(v^*)+3H\sqrt{\iota/n}$. If $n=1$, (i) holds
because $b^{\mathrm B}_z\ge16H\iota\ge H$, and (ii) because $\hat\sigma=0$.
\end{proof}

\subsection{Reference trajectories: the subsampled estimator}\label{app:traj}

Given the first $n$ steps of the stream (D2), let $M=\lfloor n/(4H)\rfloor$. A segment runs from step $0$ or a reset to the next reset. If the first $n$
steps contain fewer than $2M$ complete segments, output $\pref$. Otherwise give the first $2M$ complete segments alternately to an auxiliary
half (even segments) and a main half (odd segments), with counts $S^{\mathrm{aux}}_z$ and $S^{\mathrm{main}}_z$. With $\bar n=\lceil2MH\rceil$, $\iota=\log(8KH\bar n^2/\delta)$
and $\eta=H\iota$, let $\tilde n_z=\min\{\bar n,\lfloor\max\{S^{\mathrm{aux}}_z-16\sqrt{\eta S^{\mathrm{aux}}_z}-16\eta,0\}\rfloor\}$. If $S^{\mathrm{main}}_z\ge\tilde n_z$ for every $z$, run PSVI on the
first $\tilde n_z$ main-half samples at each pair; otherwise output $\pref$.

\begin{lemma}[two tapes]\label{lem:tape}
For each half $\mathsf h$ and pair $z$ let $(R^{\mathsf h}_{z,i},s'^{\mathsf h}_{z,i})_{i\ge1}$ be i.i.d.\ with the law of one transition at $z$, and for each segment $j$
let $U_j$ be an i.i.d.\ sequence of uniform variables; all independent. Generate segments $0,1,\dots,2M-1$ in order: segment $j$ belongs to
the auxiliary half if $j$ is even and to the main half otherwise, starts at a draw from $\rho$, and at each step at state $s$ draws
$a\sim\pref(\cdot|s)$, takes the next unread entry of its half's tape at $(s,a)$ as its reward and next state, and draws its reset coin, all
from $U_j$. These segments have the law of the first $2M$ segments of the stream (D2), extended indefinitely. The auxiliary half is a
function of the auxiliary tapes and of $(U_j)_{j\text{ even}}$, so the main tapes are independent of it.
\end{lemma}
\begin{proof}
Both constructions generate the same sequence of steps, so it suffices that at every step the conditional law of the next action,
transition and coin, given everything generated so far, is the same. The action and the coin are fresh independent draws in both. In
the tape construction the entry read at a step at pair $z$ in half $\mathsf h$ is entry number $k=1+$(number of earlier reads of that tape), and $k$
and the event that this step is at $z$ are functions of the past. The past depends on the tape $(\mathsf h,z)$ only through its first $k-1$
entries, which are independent of entry $k$, so given the past entry $k$ has the law of one transition at $z$; in (D2) this holds by
assumption. Induction over the steps gives equality of the joint laws.
\end{proof}

\begin{lemma}[length and counts]\label{lem:counts}
(i) The reset coins are i.i.d.\ Bernoulli$(1-\gamma)$, so the first $n$ steps contain at least $2M$ complete segments except with probability
$e^{-M/2}$. (ii) The number $X$ of visits of one segment to a pair $z$ satisfies $\E X=H\dref(z)$ and $\E X^k\le k!\,H^{k-1}\E X$ for $k\ge2$. (iii) In the
construction of Lemma~\ref{lem:tape}, with probability at least $1-\delta$, simultaneously for every $z$: $S^{\mathrm{main}}_z\ge\tilde n_z$, and $\tilde n_z\ge\frac18MH\dref(z)$
whenever $MH\dref(z)\ge4096\eta$.
\end{lemma}
\begin{proof}
(i) The $k$-th segment is complete within the first $n$ steps exactly when there are at least $k$ resets among them; their number is
$\mathrm{Bin}(n,1/H)$ with mean $n/H\ge4M$, and Chernoff's inequality gives $\Prob(\mathrm{Bin}(n,1/H)<2M)\le\Prob(\mathrm{Bin}(n,1/H)<\frac n{2H})\le e^{-n/(8H)}\le e^{-M/2}$.
(ii) A segment is a killed path of $\pref$ from $\rho$, so $\E X=H\dref(z)$ (Lemma~\ref{lem:occ}). Pathwise, $X^k\le k\sum_{\text{visits }t}(\text{visits from }t\text{ on})^{k-1}$; at
a visit the number of remaining visits is at most the remaining length of the segment, which given the past is geometric with mean $H$ and
$j$-th moment at most $j!H^j$, and the strong Markov property gives the claim. (iii) Let $\mu_z=MH\dref(z)$. By (ii), for each half ($M$ i.i.d.\
segments) and $|\lambda|<1/H$, $\log\E e^{\lambda(S-\mu_z)}\le H\mu_z\lambda^2/(1-H|\lambda|)$, so $|S-\mu_z|\le2\sqrt{H\mu_zt}+Ht$ except with probability $2e^{-t}$. With
$t=\log(4K/\delta)\le\iota$ and a union bound over the two halves and the $K$ pairs, both counts lie in $[\mu_z-2\sqrt{\eta\mu_z}-\eta,\ \mu_z+2\sqrt{\eta\mu_z}+\eta]$.
Expressing $\mu_z$ through the auxiliary count by $\sqrt{\mu_z}\le\sqrt{S^{\mathrm{aux}}_z+2\eta}+\sqrt \eta$ gives $S^{\mathrm{aux}}_z-S^{\mathrm{main}}_z\le4\sqrt{\eta S^{\mathrm{aux}}_z}+12\eta$, so
$S^{\mathrm{main}}_z\ge\tilde n_z$. If $\mu_z\ge4096\eta$, then $S^{\mathrm{aux}}_z\ge0.968\mu_z$ and $16\sqrt{\eta S^{\mathrm{aux}}_z}+16\eta\le0.258\mu_z$, so $\tilde n_z\ge0.71\mu_z-1\ge\mu_z/8$; the cap $\bar n$
does not bind because $\mu_z\le MH<\bar n$.
\end{proof}

\begin{proposition}\label{prop:subsample}
Let $n\ge68H$ and let PSVI use $b^{\mathrm H}$ or $b^{\mathrm B}$. With probability at least $1-2\delta-e^{-M/2}$, the subsampled estimator is PSVI on a fixed
design of the kind of Section~\ref{app:events} on which the conclusion of Proposition~\ref{prop:eventH} or~\ref{prop:eventB} holds, with counts $\tilde n_z\ge n'\dref(z)/2$
for every $z$ with $n\dref(z)\ge17408H\iota$, where $n'=MH/4\ge n/17$.
\end{proposition}
\begin{proof}
Let $F$ be the estimator applied to the first $2M$ segments of the infinite stream. On the event of Lemma~\ref{lem:counts}(i) the estimator equals $F$,
so it suffices that $F$ has the stated properties except with probability $2\delta$; no conditioning on the length event is used. By
Lemma~\ref{lem:tape}, $F$ has its law in the tape construction, where the main tapes are independent of the auxiliary half, which determines
$\tilde n$. Given $\tilde n$, the first $\tilde n_z$ entries of the main tapes are a fixed design (defined whether or not they are observed), so
Proposition~\ref{prop:eventH} or~\ref{prop:eventB} applies to it except with probability $\delta$; on the event of Lemma~\ref{lem:counts}(iii), which fails with
probability at most $\delta$, these are the data PSVI uses. Finally $MH\ge\frac4{17}n$ for $n\ge68H$, so $n\dref(z)\ge17408H\iota$ gives $MH\dref(z)\ge4096\eta$ and
$\tilde n_z\ge MH\dref(z)/8=n'\dref(z)/2$, and $n'=\frac14MH\ge\frac n{16}-\frac H4\ge\frac n{17}$.
\end{proof}

\subsection{Proof of Theorem~\ref{thm:h4}}

Let $E$ be a set of pairs with $n_z\ge\max\{n\mu(z)/2,\ c_0\iota\}$ for $z\in E$, $c_0\ge1$, and suppose $\mathcal G$ holds for PSVI with $b^{\mathrm H}$. We show, with $\cu$
depending only on $c_0$,
\[
\SubOpt_\rho(\hat\pi)\le \cu \min\Big\{\sqrt{\frac{H^4C_\Sigma(E)\iota}n}+\frac{H^2C_\Sigma(E)\iota}n,\ \frac{H^4C_\Sigma(E)\iota\Lambda}{\beta n}\Big\}+4H^2d^*(E^c),
\]
where $\Lambda=2+\log^+(n/(2c_0\iota C_\Sigma(E)))$.
Let $f(x)=\min\{x^2/\beta,x\}$, which is nondecreasing with $f(x+y)\le2f(x)+2f(y)$. Split $Z=Z_E+Z_{E^c}$ by whether $z_t\in E$. By
Proposition~\ref{prop:perturbed}, $\zeta\le H$, so $\E_*Z_{E^c}\le H\sum_{z\notin E}D^*(z)=H^2d^*(E^c)$. Lemma~\ref{lem:pathpess} gives $\SubOpt\le4\E_*Z_E^2/\beta+4H^2d^*(E^c)$ and
$\SubOpt\le\E_*Z\le\E_*Z_E+H^2d^*(E^c)$. On $E$, $\zeta\le2b^{\mathrm H}$, so $Z_E\le c_bH(\sum_{t\in E}\sqrt{\iota/n_{z_t}}+2Y)$, where $t\in E$ means $t\le T$ and $z_t\in E$, and
$Y=\sum_{t\in E}\iota/n_{z_t}$; on $E$, $\iota/n_z\le\bar w:=1/c_0$, and $\E_*Y=\sum_{z\in E}D^*(z)\iota/n_z\le\bar Y:=2H\iota C_\Sigma(E)/n$.

\emph{Slow branch.} By the Cauchy--Schwarz inequality with $\sum_zD^*(z)=H$, $\E_*\sum_{t\in E}\sqrt{\iota/n_{z_t}}=\sum_{z\in E}D^*(z)\sqrt{\iota/n_z}\le\sqrt{H\,\E_*Y}$, so
$\E_*Z_E\le c_b\sqrt{2H^4\iota C_\Sigma(E)/n}+4c_bH^2\iota C_\Sigma(E)/n$.

\emph{Fast branch.} Pathwise Cauchy--Schwarz gives $(\sum_{t\in E}\sqrt{\iota/n_{z_t}})^2\le NY$, so $Z_E^2\le2c_b^2H^2(NY+4Y^2)$. Put $y_t=\mathbf 1\{z_t\in E\}\iota/n_{z_t}$ and
$\alpha_t=\E_*[y_t\mathbf 1\{T\ge t\}]$, so $0\le\alpha_t\le\gamma^t\bar w$ and $\sum_t\alpha_t=\E_*Y$. Given the path up to time $t$ on $\{T\ge t\}$ the remaining length is
geometric and independent of the past, so $\E_*[NY]=H\E_*Y+\sum_tt\alpha_t$. With $t_0$ defined by $\gamma^{t_0}H\bar w=\E_*Y$,
$\sum_t(t-t_0)\alpha_t\le\sum_{t>t_0}(t-t_0)\gamma^t\bar w\le\gamma^{t_0}\bar wH^2=H\E_*Y$ and $t_0\le H\log(H\bar w/\E_*Y)$, so $\E_*[NY]\le H\E_*Y(2+\log(H\bar w/\E_*Y))$.
Likewise $\E_*Y^2\le2H\bar w\E_*Y$. Hence $\E_*Z_E^2\le2c_b^2H^3\E_*Y(2+\log(H\bar w/\E_*Y)+8\bar w)$. The map $x\mapsto x(2+\log(a/x))$ is increasing on
$(0,a]$ and $\E_*Y\le H\bar w$, so $\E_*Y$ may be replaced by $\min\{\bar Y,H\bar w\}$; since $H\bar w/\bar Y=n/(2c_0\iota C_\Sigma(E))$, this gives
$\E_*Z_E^2\le \cu H^4C_\Sigma(E)\iota\Lambda/n$.

Under (D1), the count event $\{n_z\ge n\mu(z)/2$ for every $z$ with $n\mu(z)\ge8\iota\}$ fails with probability at most $\sum_ze^{-n\mu(z)/8}\le Ke^{-\iota}\le\delta$ by
Chernoff's inequality, and $\mathcal G$ fails with probability at most $\delta$ (Proposition~\ref{prop:eventH}); on both, every $E$ in the statement has
$n_z\ge4\iota$ on $E$, which is part (a) with $c_0=4$ (then $2c_0=8$ in $\Lambda$). Part (b) follows from Proposition~\ref{prop:subsample} in the same way, with $n'$ and $\mu=\dref$.

\subsection{Proof of Theorem~\ref{thm:h3}}

Suppose $\mathcal G$ and the variance event of Proposition~\ref{prop:eventB}(ii) hold for PSVI with $b^{\mathrm B}$. By
Proposition~\ref{prop:perturbed}, $0\le\hat Q\le Q^*\le H$ and $0\le\hat V\le V^*\le H$, so with $W=V^*-\hat V$ and $u=Q^*-\hat Q=\zeta+\gamma PW$ we have $0\le W,u\le H$. Put
\[
g(s)=\Var_{\pi^*(\cdot|s)}u(s,\cdot),\qquad c(s)=\beta\KL(\pi^*(\cdot|s)\|\hat\pi(\cdot|s)),\qquad k_*(s)=\beta\KL(\pi^*(\cdot|s)\|\pref(\cdot|s)),
\]
and $\Gamma=\sum_{t\le T}g(s_t)$ along the killed chain of $\pi^*$; $\E_s$ is that chain started at $s$. Let $x=H/\beta$ and
\[
h_-(x)=\int_0^1(1-t)e^{tx}dt=\frac{e^x-1-x}{x^2},\qquad h_+(x)=\int_0^1te^{tx}dt=\frac{(x-1)e^x+1}{x^2},
\]
and $\alpha=2xh_-(x)$. Both integrals increase in $x$, so $\alpha$ decreases in $\beta$; $\alpha<1$ if and only if $e^x<1+\frac32x$, that is, if and only if $\beta/H$ exceeds the
root $c_\triangle\approx1.31115$ of $e^{1/c}=1+3/(2c)$.

\begin{lemma}[local tilt]\label{lem:loctilt}
Let $p$ be a distribution on a finite set, $u$ a function on it with $\max u-\min u\le x\beta$, and $\hat p\propto pe^{-u/\beta}$. Then
$\beta\KL(\hat p\|p)\le h_+(x)\Var_p(u)/\beta$ and $\beta\KL(p\|\hat p)\le h_-(x)\Var_p(u)/\beta$.
\end{lemma}
\begin{proof}
Put $\chi=u/\beta$, $\pi_\lambda\propto pe^{-\lambda\chi}$ and $\Psi(\lambda)=\log\E_pe^{-\lambda\chi}$. Then $\KL(\pi_\lambda\|p)=\lambda\Psi'(\lambda)-\Psi(\lambda)$, which vanishes at $0$
with derivative $\lambda\Var_{\pi_\lambda}(\chi)$, and $\KL(p\|\pi_\lambda)=\lambda\E_p\chi+\Psi(\lambda)$, which vanishes at $0$ with first derivative $0$ and
second derivative $\Var_{\pi_\lambda}(\chi)$. Since $\chi$ takes values in an interval of length at most $x$, $\pi_\lambda/p\le e^{\lambda x}$, so
$\Var_{\pi_\lambda}(\chi)\le\E_{\pi_\lambda}(\chi-\E_p\chi)^2\le e^{\lambda x}\Var_p(\chi)$. Integrating, $\KL(\hat p\|p)\le h_+(x)\Var_p(\chi)$ and $\KL(p\|\hat p)\le h_-(x)\Var_p(\chi)$.
\end{proof}

Since $u\in[0,H]$ and $\hat\pi\propto\pi^*e^{-u/\beta}$, Lemma~\ref{lem:loctilt} gives $c(s)\le h_-(x)g(s)/\beta$ and $\beta\KL(\hat\pi\|\pi^*)(s)\le h_+(x)g(s)/\beta$ at every state.

\begin{lemma}[pathwise total variance]\label{lem:ptv}
Let $\phi:\cZ\to[0,\infty)$ be such that the expected remaining sum of $\phi$ along the killed chain of $\pi^*$ from any pair is at most $A>0$, and
let $w:\cZ\to[0,w_{\max}]$. With $X=\sum_{t\le T}\phi(z_t)$ and $Y=\sum_{t\le T}w(z_t)$, if $\E_*Y>0$ then
\[
\E_*[XY]\ \le\ 3A\,\E_*Y\,\big(1+\log(w_{\max}H/\E_*Y)\big).
\]
\end{lemma}
\begin{proof}
The argument of Lemma~\ref{lem:khas} gives $\E_*e^{X/(2A)}\le2$. Let $\mathbb Q=(Y/\E_*Y)\Prob^*$ and $L=\log(w_{\max}H/\E_*Y)$, which is nonnegative because
$Y\le w_{\max}N$ and $\E_*N=H$. The Donsker--Varadhan inequality gives $\E_{\mathbb Q}X\le2A(\log2+\KL(\mathbb Q\|\Prob^*))$, and $\E_*[XY]=\E_*Y\cdot\E_{\mathbb Q}X$. By
$Y\le w_{\max}N$ and Jensen's inequality, $\KL(\mathbb Q\|\Prob^*)=\E_{\mathbb Q}\log(Y/\E_*Y)\le\log(w_{\max}/\E_*Y)+\log\E_{\mathbb Q}N$, and
$\E_{\mathbb Q}N=\E_*[NY]/\E_*Y\le H(2+L)$ by the argument for $\E_*[NY]$ in the proof of Theorem~\ref{thm:h4}. So $\KL(\mathbb Q\|\Prob^*)\le L+\log(2+L)\le\log2+\frac32L$,
and $\E_{\mathbb Q}X\le A(4\log2+3L)\le3A(1+L)$.
\end{proof}

\begin{lemma}[energy bounds]\label{lem:energy}
Suppose $\alpha<1$. Then for every state $s$: (i) $\E_s\Gamma\le\bar\Gamma:=8H^2/(1-\alpha)^2$; (ii) $\E_s\sum_{t\le T}\gamma^2\Var_P\hat V(z_t)\le2H^2+\alpha\bar\Gamma$, and
the same sum started from any pair is at most $\bar\Gamma_{\hat V}:=\frac94H^2+\alpha\bar\Gamma$. For $\beta\ge8H$, $\alpha\le0.1304$, $\bar\Gamma\le10.58H^2$ and
$\bar\Gamma_{\hat V}\le3.63H^2$; for $\beta\ge1.4H$, $\alpha\le0.9197$, $\bar\Gamma\le1239H^2$ and $\bar\Gamma_{\hat V}\le1142H^2$.
\end{lemma}
\begin{proof}
For $f\in\{V^*,\hat V\}$ and $q=r+\gamma Pf$ (if $f=V^*$) or $q=\hat Q$ (if $f=\hat V$): in both cases $0\le q\le r+\gamma Pf$ and $f\in[0,H]$, so
$q^2\le r^2+2r\gamma Pf+\gamma^2(Pf)^2\le2Hr+\gamma P(f^2)-\gamma^2\Var_Pf$, using $r\le1$, $1+2\gamma H\le2H$ and $\gamma^2(Pf)^2\le\gamma P(f^2)-\gamma^2\Var_Pf$. Next,
$\pi^*Q^*=V^*+k_*\ge V^*$, and the Gibbs variational identity gives $\pi^*\hat Q=\hat V+k_*-c$, so $(\pi^*\hat Q)^2\ge\hat V^2-2Hc$ (if $\hat V\ge c$ this is
$(\hat V-c)^2\ge\hat V^2-2\hat Vc$; otherwise $\hat V^2<Hc$). Hence
\[
\Var_{\pi^*}q+\gamma^2\pi^*\Var_Pf\ \le\ 2H\pi^*r+\gamma P_{\pi^*}(f^2)-f^2+2Hc\,\mathbf 1\{f=\hat V\}.
\]
Along the killed chain the terms $\gamma P_{\pi^*}(f^2)(s_t)-f(s_t)^2$ telescope in expectation to $-f(s)^2\le0$, and $\E_s\sum_t\pi^*r(s_t)\le H$ (this is the
total-variance argument of \citet{azar2013minimax} in soft form). With
$J_s(q,f)=\E_s\sum_{t\le T}[\Var_{\pi^*}q(s_t,\cdot)+\gamma^2\Var_Pf(z_t)]$ this gives $J_s(Q^*,V^*)\le2H^2$ and, by Lemma~\ref{lem:loctilt},
$J_s(\hat Q,\hat V)\le2H^2+2H\E_s\sum_tc(s_t)\le2H^2+\alpha\E_s\Gamma$. Both are taken under the path law of $\pi^*$, so $\sqrt{J_s}$ is a seminorm, and
$\E_s\Gamma\le J_s(Q^*-\hat Q,V^*-\hat V)$ gives $\sqrt{\E_s\Gamma}\le\sqrt2H+\sqrt{2H^2+\alpha\E_s\Gamma}$, where $\E_s\Gamma\le H^3/4$ is finite. If $\sqrt{\E_s\Gamma}>\sqrt2H$,
subtracting $\sqrt2H$ and squaring gives $(1-\alpha)\E_s\Gamma\le2\sqrt2H\sqrt{\E_s\Gamma}$, so $\E_s\Gamma\le\bar\Gamma$; otherwise $\E_s\Gamma\le2H^2\le\bar\Gamma$. This is (i), and
(ii) follows from $J_s(\hat Q,\hat V)\le2H^2+\alpha\bar\Gamma$; from a forced first pair add $\gamma^2\Var_P\hat V(z)\le H^2/4$. The values at $8H$ and $1.4H$ follow from
the formulas, since $\alpha$ decreases in $\beta$.
\end{proof}

\begin{lemma}[action-variance pessimism]\label{lem:avp}
Suppose $\alpha<1$ and $\psi:=h_+(x)\bar\Gamma/\beta^2<1$. Then $\SubOpt_\rho(\hat\pi)\le\dfrac{h_+(x)}{(1-\sqrt\psi)^2}\,\dfrac{\E_*\Gamma}\beta$. Both conditions hold for $\beta\ge3.33H$,
and the constant $h_+(x)/(1-\sqrt\psi)^2$ decreases in $\beta$; it is at most $40.8$ at $\beta=3.6H$ and at most $1.109$ for $\beta\ge8H$.
\end{lemma}
\begin{proof}
$\E_s\Gamma^j\le j!\bar\Gamma^{j-1}\E_s\Gamma$ for every $s$ and $j\ge1$: with $\Gamma_{\ge t}=\sum_{t\le k\le T}g(s_k)$, $\Gamma^j\le j\sum_tg(s_t)\Gamma_{\ge t}^{j-1}$, and given $s_t$
(on $\{T\ge t\}$) $\Gamma_{\ge t}$ has the law of $\Gamma$ started at $s_t$, so induction and Lemma~\ref{lem:energy}(i) give $\E[\Gamma_{\ge t}^{j-1}|s_t]\le(j-1)!\bar\Gamma^{j-1}$. Summing the
series, $\log\E_*e^{\theta\Gamma}\le\theta\E_*\Gamma/(1-\theta\bar\Gamma)$ for $\theta\bar\Gamma<1$. By Lemma~\ref{lem:pathkl}, the chain rule and Lemma~\ref{lem:loctilt}, with $h_+=h_+(x)$,
$\SubOpt=\beta\hat\E\sum_{t\le T}\KL(\hat\pi\|\pi^*)(s_t)\le h_+\hat\E\Gamma/\beta$, an expectation under $\hat\pi$'s path law. The Donsker--Varadhan inequality gives
$\theta\hat\E\Gamma\le\KL(\hat\Prob\|\Prob^*)+\log\E_*e^{\theta\Gamma}$ with $\KL(\hat\Prob\|\Prob^*)=\SubOpt/\beta$, so $(\theta\beta^2-h_+)\SubOpt\le h_+\beta\theta\E_*\Gamma/(1-\theta\bar\Gamma)$ for
$h_+/\beta^2<\theta<1/\bar\Gamma$, and $\theta=\sqrt{h_+}/(\beta\sqrt{\bar\Gamma})$ gives the claim. The numerical statements follow from the formulas, since $h_+(x)$, $\alpha$ and
$\bar\Gamma/H^2$ decrease in $\beta$ and so does $\psi$.
\end{proof}

\begin{proof}[Proof of Theorem~\ref{thm:h3}]
Let $\beta\ge1.4H$ and let $E$ be a set of pairs with $n_z\ge\max\{n\mu(z)/2,8\iota\}$ on $E$; we show
$\SubOpt\le \cu [H^3C_\Sigma(E)\iota L_n/(\beta n)+H^3d^*(E^c)/\beta]$. Write $w_z=\iota/n_z$, so $w_z\le\frac18$ on $E$, and $Y=\sum_{t\le T:z_t\in E}w_{z_t}$.

\emph{The split.} The proof of Proposition~\ref{prop:eventB} shows that on its event $\hat\sigma_z(\hat V)\le\sigma_z+3H\sqrt{w_z}$, where $\sigma_z$ is the standard deviation of
$R+\gamma\hat V(s')$ under the true law at $z$ with $\hat V$ held fixed, and $\sigma_z\le\frac12+\sqrt{\gamma^2\Var_P\hat V(z)}$ by the triangle inequality for standard
deviations. On $E$, $\zeta\le2b^{\mathrm B}=\max\{8\hat\sigma\sqrt w,32Hw\}$, so
\[
\zeta\ \le\ 4\sqrt w+8\sqrt{\gamma^2\Var_P\hat V\,w}+32Hw\qquad\text{on }E.
\]
Put $\zeta_1=\min\{\zeta,4\sqrt w+8\sqrt{\gamma^2\Var_P\hat V\,w}\}$ and $\zeta_2=\zeta-\zeta_1$ on $E$, and $\zeta_1=0$, $\zeta_2=\zeta$ off $E$. Then $\zeta_1,\zeta_2\ge0$, $\zeta_2\le32Hw$ on
$E$ and $\zeta_2\le H$ off $E$. Write $Z_i=\sum_{t\le T}\zeta_i(z_t)$, $A_1=\E_*Z_1^2$ and $D_2=\E_*Z_2$; then $D_2\le32H\E_*Y+H^2d^*(E^c)$.

\emph{The two path sums.} Suppose $d^*(E)\ge\frac12$. Since $n_z\le n$, $\E_*Y\ge H\iota/(2n)$ (in part (b) only $\tilde n_z\le\bar n\le8n'+1$ holds, so $\E_*Y\ge H\iota/(18n')$,
which changes only the absolute constants below), and $L=\log(H/(8\E_*Y))$ satisfies $2+L\le L_n$. With $S_w=\sum_{t\le T:z_t\in E}\sqrt{w_{z_t}}$ and
$S_v=\sum_{t\le T:z_t\in E}\sqrt{\gamma^2\Var_P\hat V(z_t)w_{z_t}}$, $Z_1\le4S_w+8S_v$, so $\sqrt{A_1}\le4\sqrt{\E_*S_w^2}+8\sqrt{\E_*S_v^2}$. The pathwise Cauchy--Schwarz inequality
gives $S_w^2\le NY$ and $S_v^2\le X_{\hat V}Y$ with $X_{\hat V}=\sum_{t\le T}\gamma^2\Var_P\hat V(z_t)$. As in the proof of Theorem~\ref{thm:h4}, $\E_*[NY]\le H\E_*Y(2+L)$, and
Lemma~\ref{lem:ptv} with Lemma~\ref{lem:energy}(ii) gives $\E_*[X_{\hat V}Y]\le3\bar\Gamma_{\hat V}\E_*Y(1+L)$. With $H\ge2$,
\[
A_1\ \le\ a_1H^2\E_*Y\,L_n,\qquad a_1=\big(2\sqrt2+8\sqrt{3\bar\Gamma_{\hat V}/H^2}\big)^2 .
\]

\emph{Case $\beta\ge3.6H$: the energy.} Fix $\lambda>0$. Since $u=\zeta+\gamma PW$, Young's inequality gives
$g\le(1+\lambda)\Var_{\pi^*}\zeta+(1+\lambda^{-1})\gamma^2\Var_{\pi^*}(PW)\le(1+\lambda)\pi^*(\zeta^2)+(1+\lambda^{-1})\gamma^2\Var_{\pi^*}(PW)$. With $W_t=W(s_t)$ and $W_{T+1}=0$, the law of total
variance gives $\gamma^2\Var_{\pi^*}(PW)(s_t)\le\Var(W_{t+1}|s_t)\le\E[(W_{t+1}-W_t)^2|s_t]$. The Gibbs identity of Lemma~\ref{lem:energy}'s proof gives
$W=\pi^*(\zeta+\gamma PW)-c$, so $\E[W_{t+1}-W_t|s_t]=c(s_t)-\pi^*\zeta(s_t)$; expanding $W_{T+1}^2-W_0^2=\sum_t[(W_{t+1}-W_t)^2+2W_t(W_{t+1}-W_t)]$ and using $W,c\ge0$,
$\E_*\sum_t(W_{t+1}-W_t)^2\le2\E_*\sum_tW_t\zeta(z_t)$. Hence $\E_*\Gamma\le(1+\lambda)\E_*\sum_t\zeta(z_t)^2+2(1+\lambda^{-1})\E_*\sum_tW_t\zeta(z_t)$.

\emph{The cross term.} $\hat V$ is the soft-optimal value for $r-\zeta$, so $\hat V\ge V^{\pi^*}_{r-\zeta}$ and $W_t\le\E[Z_{\ge t}|\mathcal F_t]$; also $W_t\le H$. With the martingales
$M_{i,t}=\E[Z_i|\mathcal F_t]$, $W_t\le M_{1,t}+\min\{H,M_{2,t}\}$, so $W_{\max}=\sup_tW_t\le\sup_tM_{1,t}+\min\{H,\sup_tM_{2,t}\}$. Doob's $L^2$ inequality gives
$\E(\sup_tM_{1,t})^2\le4A_1$, and the maximal inequality $\Prob(\sup_tM_{2,t}>y)\le D_2/y$ gives $\E\min\{H,\sup_tM_{2,t}\}^2\le2HD_2$; so $\E W_{\max}^2\le8A_1+4HD_2$. Then
\[
\E_*\sum_tW_t\zeta(z_t)\ \le\ \E_*[W_{\max}Z_1]+H\,\E_*Z_2\ \le\ \sqrt{A_1(8A_1+4HD_2)}+HD_2\ \le\ 3A_1+2HD_2 .
\]
On $E$, $\zeta^2\le128H^2w$ (from $\hat\sigma\le H/2$ and $w\le\frac18$), and off $E$, $\zeta^2\le H\zeta$; so $\E_*\sum_t\zeta(z_t)^2\le128H^2\E_*Y+H^3d^*(E^c)$, and
\[
\E_*\Gamma\ \le\ F(\lambda)H^2\E_*Y\,L_n+\big(5+\lambda+4\lambda^{-1}\big)H^3d^*(E^c),\qquad F(\lambda)=64\big(2+\lambda+\lambda^{-1}\big)+6\big(1+\lambda^{-1}\big)a_1,
\]
using $L_n\ge2$ (the coefficient of $H^2\E_*Y$ before that step is $128(2+\lambda+\lambda^{-1})$). Lemma~\ref{lem:avp} and $\E_*Y\le2H\iota C_\Sigma(E)/n$ give the claim. If $d^*(E)<\frac12$, Lemmas~\ref{lem:avp} and~\ref{lem:energy}(i) give
$\SubOpt\le40.8\,\bar\Gamma/\beta\le82\,\bar\Gamma d^*(E^c)/\beta$.

\emph{Case $1.4H\le\beta\le3.6H$: the path-space route.} Lemma~\ref{lem:pathpess} and $f(y+y')\le2f(y)+2f(y')$, $f(y)=\min\{y^2/\beta,y\}$, give
$\SubOpt\le4\E_*Z_1^2/\beta+4\E_*Z_2=4A_1/\beta+4D_2$, so for $d^*(E)\ge\frac12$
\[
\SubOpt\ \le\ 8a_1\frac{H^3C_\Sigma(E)\,\iota\,L_n}{\beta n}+256\,\frac{H^2C_\Sigma(E)\,\iota}n+4H^2d^*(E^c),
\]
and $H^2\le3.6H^3/\beta$ puts the last two terms in the form of the claim. If $d^*(E)<\frac12$, $\SubOpt\le\E_*Z\le H^2\le7.2H^3d^*(E^c)/\beta$.

\emph{Conclusion.} In both cases $a_1$ and the other constants are bounded for $\beta\ge1.4H$ by Lemma~\ref{lem:energy}. Under (D1), the count event of the
proof of Theorem~\ref{thm:h4} (for $n\mu(z)\ge16\iota$ it gives $n_z\ge n\mu(z)/2\ge8\iota$) and Proposition~\ref{prop:eventB} hold together with probability at least
$1-2\delta$, which gives part (a); for the last claim take $E=\{z:n\mu(z)\ge16\iota\}$, for which $d^*(E^c)\le\frac{16\iota}nC_\Sigma$. Part (b) follows from
Proposition~\ref{prop:subsample}, since $\tilde n_z\ge n'\dref(z)/2$ and $n'\dref(z)/2\ge512H\iota\ge8\iota$ on the stated pairs.
\end{proof}

\begin{remark}[constants and the range of temperatures]\label{rem:h3range}
The constant of Theorem~\ref{thm:h3} depends on $\beta/H$ through $\alpha$ and $\psi$. With $c_\triangle\approx1.31115$ as above, the path-space route works for every
$\beta/H\in(c_\triangle,3.6]$, with a constant of order $(1-\alpha)^{-2}$ that grows without bound as $\beta/H$ decreases to $c_\triangle$; below $c_\triangle$ the energy argument gives no
bound on $\E_s\Gamma$, and Proposition~\ref{prop:burnin} needs a burn-in. The path-space route also holds for $\beta>3.6H$, with main constant $8a_1$ (about
$6833$ at $\beta=8H$), but there its term $256H^2C_\Sigma(E)\iota/n$ does not decrease with $\beta$. Appendix~\ref{app:checks} evaluates Lemmas~\ref{lem:loctilt}--\ref{lem:avp} numerically on small random instances.
\end{remark}

\subsection{Proof of Corollary~\ref{cor:h3}}\label{app:cor}

The claim is the last statement of Theorem~\ref{thm:h3}(a), since $C_\Sigma\le KC^*(\mu)$. For the expected loss: at every state
$\beta\KL(\hat\pi(\cdot|s)\|\pref(\cdot|s))\le\max\hat Q-\min\hat Q\le H$, so $V^{\hat\pi}\ge-H^2$ and $\SubOpt_\rho(\hat\pi)\le H+H^2$ for every dataset. With
$\delta=\epsilon/(2(H+H^2))$ this gives $\E[\SubOpt_\rho(\hat\pi)]\le\epsilon+2\delta(H+H^2)=2\epsilon$, and $\iota$ grows by $\log(2(H+H^2)/\epsilon)$.

\subsection{Proof of Corollary~\ref{cor:refdata}}\label{app:refdata}

Let $\cu\ge4$ be at least the constants of Theorems~\ref{thm:h4}(b) and~\ref{thm:h3}(b). Write $\tau=\tau_n$, so $\tau^{1+a}=17408H\iota e^{c_1H/\beta}/n$, and let
$E=\{z:d^*(z)>\tau\}$. For $z\in E$, Theorem~\ref{thm:coverage} applied to the set $\{z\}$ gives
$\dref(z)\ge e^{-c_1H/\beta}d^*(z)^{1+a}>e^{-c_1H/\beta}\tau^{1+a}=17408H\iota/n$, so $E$ is one of the sets allowed in Theorems~\ref{thm:h4}(b) and~\ref{thm:h3}(b). It also gives
$d^*(z)/\dref(z)\le e^{c_1H/\beta}d^*(z)^{-a}<e^{c_1H/\beta}\tau^{-a}$, so with $\mu=\dref$,
\[
C_\Sigma(E)\ \le\ \bar x:=Ke^{c_1H/\beta}\tau^{-a},\qquad d^*(E^c)=\sum_{z:d^*(z)\le\tau}d^*(z)\ \le\ K\tau .
\]
Since $n'\ge n/17$, $\iota\bar x/n'\le17K\iota e^{c_1H/\beta}\tau^{-a}/n=17K\tau/(17408H)=K\tau/(1024H)$. Both theorems hold on their events for every allowed set at
once, so we may use this $E$. If $E=\emptyset$, then $C_\Sigma(E)=0$ and $d^*(E^c)=1\le K\tau$, and both claims follow from the uncovered terms; assume $E\neq\emptyset$, so
$\tau<1$ and $\bar x\ge K\ge1$.

For $b^{\mathrm B}$ and $\beta\ge1.4H$, Theorem~\ref{thm:h3}(b) with $L_{n'}\le L_n$ gives
$\SubOpt_\rho(\hat\pi)\le \cu [H^3\bar x\iota L_n/(\beta n')+H^3K\tau/\beta]\le \cu [H^2L_nK\tau/(1024\beta)+H^3K\tau/\beta]$. For $b^{\mathrm H}$ we use the second branch of
Theorem~\ref{thm:h4}(b). The map $x\mapsto x(2+\log^+(n'/(8\iota x)))$ is nondecreasing on $x>0$, so $C_\Sigma(E)\Lambda\le\bar x(2+\log^+(n'/(8\iota\bar x)))\le\bar xL_n$, using
$\bar x\ge1$ and $8\iota\ge1$; hence $\SubOpt_\rho(\hat\pi)\le \cu H^4\bar x\iota L_n/(\beta n')+4H^2K\tau\le \cu H^3L_nK\tau/(1024\beta)+4H^2K\tau$. This proves both bounds.

For the last claim, $H^3+H^2L_n\le H^3L_n$ (since $H\ge2$ and $L_n\ge2$), so with $\cu \ge1$ it suffices that $\tau\le\beta\epsilon/(\cu KH^3L_n)$, that is
$n\ge17408H\iota e^{c_1H/\beta}(\cu KH^3L_n/(\beta\epsilon))^{1+a}$. For $\beta\ge1.4H$, $e^{c_1H/\beta}\le e^{2.75/1.4}\le7.2$ and $\cu ^{1+a}\le \cu ^{1+6/1.4}$, which gives the claim
with $c=17408\cdot7.2\cdot \cu ^{1+6/1.4}$.

\subsection{Every temperature after a burn-in}\label{app:burnin}

The proof of Theorem~\ref{thm:h3} uses the condition on $\beta$ in three places: the energy bounds of Lemma~\ref{lem:energy}, the exponential moment of
$\Gamma$ in Lemma~\ref{lem:avp}, and the absorption of the range term in the path-space route. For every $\beta$, a count condition on $E$ can replace all
three. Here we split the slack through $V^*$ and $W$: by Proposition~\ref{prop:eventB}(ii) and the triangle inequality for standard deviations,
$\hat\sigma_z\le\frac12+\sqrt{\Var_PV^*(z)}+\sqrt{\gamma^2\Var_PW(z)}+3H\sqrt{\iota/n_z}$, so on $E$
\[
\zeta\ \le\ c_s\Big(\sqrt{\iota\Var_PV^*/n_z}+\sqrt{\iota\gamma^2\Var_PW/n_z}+\sqrt{\iota/n_z}+H\iota/n_z\Big),\qquad c_s=72,
\]
and we put $c_{\mathrm{burn}}=128c_s^2$. The killed variance $v=\gamma P(V^{*2})-\gamma^2(PV^*)^2$ of Lemma~\ref{lem:tvbound} satisfies $v\ge\gamma\Var_PV^*$, so the remaining
sum of $\Var_PV^*$ is at most $2H^2/\gamma$ from a state and at most $H^2/4+\gamma\cdot2H^2/\gamma=2.25H^2$ from a pair.

\begin{lemma}[mixed energy]\label{lem:mixed}
Suppose $0\le\hat V\le V^*\le H$ and $\hat\pi$ is the soft-optimal policy for $r-\zeta$ with $\zeta\ge0$, and let $E$ be a set of pairs, $A=\E_*Z_E^2$ and
$D=\E_*Z_{E^c}$. Then $\E_*\sum_{t\le T}\gamma^2\Var_PW(z_t)\le6A+4HD$, with $W=V^*-\hat V$.
\end{lemma}
\begin{proof}
As in the proof of Theorem~\ref{thm:h3}, $W_t\le\E[Z_E|\mathcal F_t]+\min\{H,\E[Z_{E^c}|\mathcal F_t]\}$, so $\E(\sup_tW_t)^2\le8A+4HD$ by Doob's $L^2$ inequality and the
maximal inequality, and $\E_*\sum_tW_t\zeta(z_t)\le\E_*[\sup_tW_t\,Z_E]+HD\le3A+2HD$. The energy identity of that proof gives
$\E_*\sum_t(W_{t+1}-W_t)^2\le2\E_*\sum_tW_t\zeta(z_t)$, and $\E[(W_{t+1}-W_t)^2|z_t]\ge\Var(W_{t+1}|z_t)\ge\gamma\Var_PW(z_t)\ge\gamma^2\Var_PW(z_t)$. None of these steps uses
the value of $\beta$.
\end{proof}

\begin{proposition}[every temperature after a burn-in]\label{prop:burnin}
Let $\beta>0$ and $n\ge H$. Under (D1), with probability at least $1-2\delta$, PSVI with $b^{\mathrm B}$ satisfies, for every
$E\subseteq\{z:n\mu(z)\ge2c_{\mathrm{burn}}H\iota L_n\}$,
\[
\SubOpt_\rho(\hat\pi)\ \le\ \cu \Big[\frac{H^3C_\Sigma(E)\,\iota\,L_n}{\beta n}+\Big(H^2+\frac{H^3}\beta\Big)d^*(E^c)\Big].
\]
In particular, if $\mu\ge\mu_0>0$ on the pairs with $d^*>0$, then for each $\beta$, with probability at least $1-2\delta$,
$\SubOpt_\rho(\hat\pi)\le \cu H^3C_\Sigma\iota L_n/(\beta n)$ once $n\ge2c_{\mathrm{burn}}H\iota L_n/\mu_0$.
\end{proposition}
\begin{proof}
Work on the count event and the events of Proposition~\ref{prop:eventB} for the given $\beta$ (PSVI's fixed point, and so these events, depend on $\beta$),
which hold together with probability at least $1-2\delta$; then $n_z\ge n\mu(z)/2\ge c_{\mathrm{burn}}H\iota L_n$ on $E$, so $\bar w:=\max_E\iota/n_z\le1/(c_{\mathrm{burn}}HL_n)$. If $d^*(E)<\frac12$, Lemma~\ref{lem:pathpess} and $\zeta\le H$ give
$\SubOpt\le\E_*Z\le H^2\le2H^2d^*(E^c)$; assume $d^*(E)\ge\frac12$, so $H\iota/(2n)\le\E_*Y\le2H\iota C_\Sigma(E)/n$ with $Y=\sum_{t\in E}\iota/n_{z_t}$. The bound on
$\zeta$ above gives $Z_E\le c_s(S_1+S_2+S_3+HY)$ with $S_1,S_2,S_3$ the path sums over $E$ of the three square roots, and
$A\le4c_s^2(\E_*S_1^2+\E_*S_2^2+\E_*S_3^2+H^2\E_*Y^2)$. Lemma~\ref{lem:ptv} with the remaining sums $2.25H^2$ of $\Var_PV^*$ gives $\E_*S_1^2\le cH^2\E_*Y\,L_n$, and
$\E_*S_3^2\le\E_*[NY]\le H\E_*Y\,L_n$ as in the proof of Theorem~\ref{thm:h4}; also $\E_*Y^2\le2H\bar w\E_*Y\le2\E_*Y/(c_{\mathrm{burn}}L_n)$. For $S_2$, the pathwise Cauchy--Schwarz inequality gives
$S_2^2\le X_WY\le\bar wX_WN$ with $X_W=\sum_{t\le T}\gamma^2\Var_PW(z_t)$, and since $\gamma^2\Var_PW\le H^2/4$ the argument for $\E_*[NY]$ in the proof of
Theorem~\ref{thm:h4} gives $\E_*[X_WN]\le Hx(2+\log^+(H^3/(4x)))$ with $x=\E_*X_W$. The right side is nondecreasing in $x$, and $x\le6A+4HD$ by
Lemma~\ref{lem:mixed}; replacing $x$ by $\max\{6A+4HD,H^3/n\}$ bounds the logarithm by $\log n$, so $\E_*S_2^2\le\bar wHL_n(6A+4HD+H^3/n)\le(6A+4HD+H^3/n)/c_{\mathrm{burn}}$.
Since $4c_s^2/c_{\mathrm{burn}}=\frac1{32}$, the $S_2$ term contributes at most $(6A+4HD+H^3/n)/32$, which is absorbed: $A\le cH^2\E_*Y\,L_n+HD/6+H^3/(26n)$, and
$H^3/n\le2H^2\E_*Y$. Finally Lemma~\ref{lem:pathpess} with $f(x+y)\le2f(x)+2f(y)$, $f(x)=\min\{x^2/\beta,x\}$, gives $\SubOpt\le4A/\beta+4D$, and
$D\le H^2d^*(E^c)$, $\E_*Y\le2H\iota C_\Sigma(E)/n$ give the claim. For the last statement take $E=\{z:d^*(z)>0\}$.
\end{proof}

\subsection{Explicit constants}\label{app:constants}

Tracking the constants in the proofs above gives the following. In the proof of Theorem~\ref{thm:h4}, $c_b=\sqrt2$ and $c_0=4$ give, for
part (a),
\begin{align*}
\SubOpt_\rho(\hat\pi)\le\min\Big\{&2\sqrt{\frac{H^4C_\Sigma(E)\iota}n}+5.66\,\frac{H^2C_\Sigma(E)\iota}n+H^2d^*(E^c),\\
&64\,\frac{H^4C_\Sigma(E)\iota\Lambda}{\beta n}+4H^2d^*(E^c)\Big\}.
\end{align*}
In the proof of Theorem~\ref{thm:h3} at $\beta\ge8H$, Lemma~\ref{lem:energy} gives $\bar\Gamma_{\hat V}\le3.63H^2$, so $a_1\le(2\sqrt2+8\sqrt{10.89})^2\le854.31$ and
the choice $\lambda=9$ (close to the minimizer $\sqrt{1+6a_1/64}\approx9.0$) gives $F(9)\le6407$ and
$\E_*\Gamma\le6407H^2\E_*Y\,L_n+14.45H^3d^*(E^c)$. With the constant $1.109$ of Lemma~\ref{lem:avp} and $\E_*Y\le2H\iota C_\Sigma(E)/n$,
part (a) reads
\[
\SubOpt_\rho(\hat\pi)\le1.422\cdot10^4\,\frac{H^3C_\Sigma(E)\,\iota\,L_n}{\beta n}+23.5\,\frac{H^3}\beta\,d^*(E^c),
\]
where $1.422\cdot10^4\ge2\cdot1.109\cdot6407$, and $23.5\ge2\cdot1.109\cdot10.58$ covers the case $d^*(E)<\frac12$ (and exceeds $1.109\cdot14.45$). For $1.4H\le\beta\le3.6H$ the path-space route of the proof gives the main-term constant
$8a_1+461$, which falls from about $1.8\cdot10^6$ at $\beta=1.4H$ to $1.4\cdot10^4$ at $3.6H$. These constants were not optimized. For the same set $E$
with $d^*(E)\ge\frac12$, where $\Lambda\le L_n$, the main term of the second bound is below that of the first only when $H>\CTwentyOneHCross\,L_n/\Lambda$, so the
gain from $H^4$ to $H^3$ appears at long horizons.

To see how far the bounds are from being useful, we evaluated both, minimized over the sets $E$ they allow, on two of our instances. On the
loop family of Section~\ref{sec:experiments} ($\beta=8H$), at the simulated size $n=4\cdot10^6$ the bound of Theorem~\ref{thm:h3} is at least $\CTwentyOneLoopRatioMin$ times
the loss of the reference policy. It needs at least $n=\CTwentyOneLoopNRefMin$ to drop below that loss at any of the horizons we use, and the bound of
Theorem~\ref{thm:h4} needs at least $n=\CTwentyOneLoopTOneNRefMin$. Theorem~\ref{thm:h3} does not apply to the instance of Theorem~\ref{thm:plugin}; there the bound of Theorem~\ref{thm:h4} is at
least $\CTwentyOnePlugRatioMin$ times the reference policy's loss at the sample sizes we simulate, and drops below it at about $n=\CTwentyOnePlugNRef$. On the events of the theorems PSVI
also satisfies $\SubOpt_\rho(\hat\pi)\le V^*(\rho)\le H$, which at these sizes is smaller than the bound of Theorem~\ref{thm:h3} on the loop family and
than that of Theorem~\ref{thm:h4} on the second instance, and on the loop family the losses PSVI attains are far below the reference policy's
(Figure~\ref{fig:c19c}).

\section{The gadget and the proof of Theorem~\ref{thm:lb-design}}\label{app:lower}

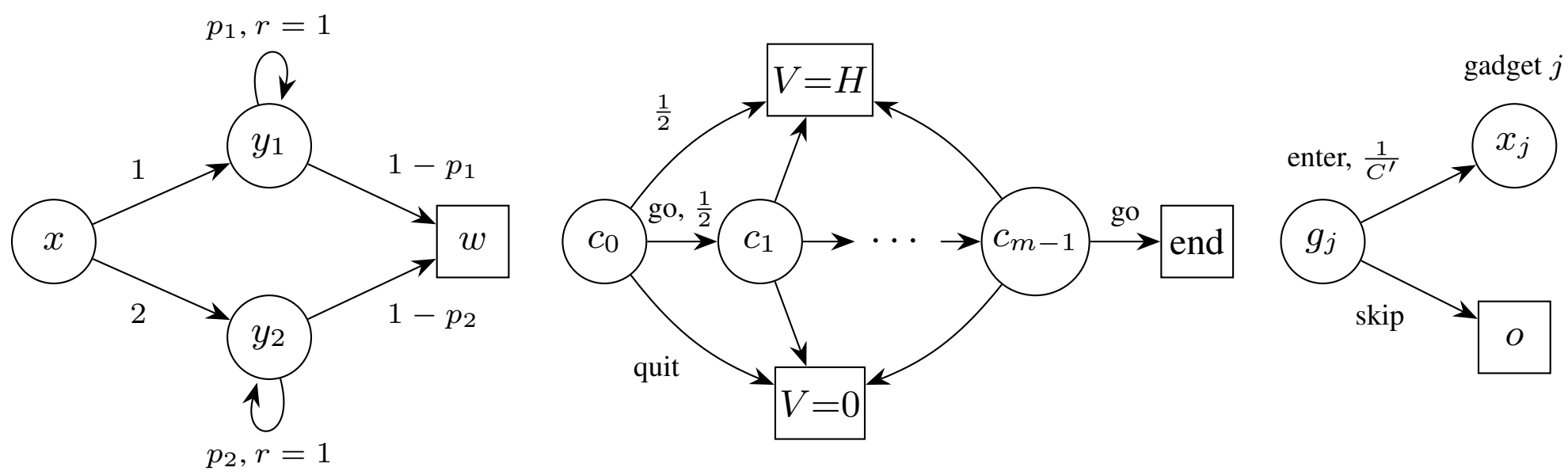
\begin{figure}[htb]
\centering
\begin{tikzpicture}[>={Stealth[length=2mm]}, st/.style={circle,draw,minimum size=7mm,inner sep=0pt,font=\small},
  ab/.style={rectangle,draw,minimum size=6mm,inner sep=1pt,font=\small}, lab/.style={font=\scriptsize}, every edge/.style={draw,->}]
\node[st] (x) at (0,0) {$x$};
\node[st] (y1) at (1.8,0.8) {$y_1$};
\node[st] (y2) at (1.8,-0.8) {$y_2$};
\node[ab] (w) at (3.5,0) {$w$};
\draw (x) edge node[lab,above left] {$1$} (y1) (x) edge node[lab,below left] {$2$} (y2);
\draw (y1) edge[loop above] node[lab,above] {$p_1$, $r=1$} (y1) (y2) edge[loop below] node[lab,below] {$p_2$, $r=1$} (y2);
\draw (y1) edge node[lab,above right] {$1-p_1$} (w) (y2) edge node[lab,below right] {$1-p_2$} (w);
\node[font=\small] at (1.75,-2.25) {(a) gadget};
\begin{scope}[xshift=4.6cm]
\node[st] (c0) at (0,0) {$c_0$};
\node[st] (c1) at (1.3,0) {$c_1$};
\node (dots) at (2.45,0) {$\cdots$};
\node[st,minimum size=9mm] (cm) at (3.6,0) {$c_{m-1}$};
\node[ab] (end) at (4.95,0) {end};
\node[ab] (hv) at (1.8,1.35) {$V{=}H$};
\node[ab] (zr) at (1.8,-1.35) {$V{=}0$};
\draw (c0) edge node[lab,above] {go, $\frac12$} (c1) (c1) edge (dots) (dots) edge (cm) (cm) edge node[lab,above] {go} (end);
\draw (c0) edge[bend left=15] node[lab,above left] {$\frac12$} (hv) (c1) edge (hv) (cm) edge[bend right=15] (hv);
\draw (c0) edge[bend right=15] node[lab,below left] {quit} (zr) (c1) edge (zr) (cm) edge[bend left=15] (zr);
\node[font=\small] at (2.45,-2.25) {(b) chain};
\end{scope}
\begin{scope}[xshift=10.6cm]
\node[st] (g) at (0,0) {$g_j$};
\node[st] (xj) at (1.6,0.8) {$x_j$};
\node[ab] (o) at (1.6,-0.8) {$o$};
\draw (g) edge node[lab,above left] {enter, $\frac1{C'}$} (xj) (g) edge node[lab,below left] {skip} (o);
\node[lab] at (1.6,1.45) {gadget $j$};
\node[font=\small] at (1.0,-2.25) {(c) gate};
\end{scope}
\end{tikzpicture}
\caption{The constructions of the lower bounds. (a) The two-armed gadget: the reference policy is uniform at $x$, and the loop
probabilities $p_{1,2}=p_0\pm\vartheta\Delta$ carry the sign $\vartheta$; $w$ is absorbing with reward $0$. (b) The chain of Lemma~\ref{lem:chain}: go
(reference probability $\frac12$) moves on with probability $\frac12$ and otherwise to an absorbing state of value $H$; quit moves to an absorbing
zero-reward state. Theorem~\ref{thm:power} puts the gadgets after the end state. (c) A gate of Theorem~\ref{thm:gate} in front of each gadget,
$C'=C/2$; $o$ is absorbing with reward $0$.}
\label{fig:constructions}
\end{figure}

Rewards are known and deterministic in every construction of Appendices~\ref{app:lower}, \ref{app:reference}
and~\ref{app:plugin}, so the observed rewards carry no information about the unknown parameter. Figure~\ref{fig:constructions} shows the constructions.

\paragraph{The two-armed gadget \citep{liu2026sample}.} Fix $\gamma\ge4/7$, $p_0=(4\gamma-1)/(3\gamma)$ and
$0\le\Delta\le\Delta_{\max}:=(1-p_0)/2=1/(6\gamma H)$. A gadget has a decision state $x$ with two actions and reference policy
uniform on them; action $i\in\{1,2\}$ moves to a loop state $y_i$ with reward $0$; at $y_i$ the reward is $1$, the next state is
$y_i$ with probability $p_i$ and an absorbing zero-reward state $w$ otherwise; $p_1=p_0+\vartheta\Delta$ and $p_2=p_0-\vartheta\Delta$ for a
sign $\vartheta\in\{\pm1\}$. Only the transitions at $y_1,y_2$ depend on $\vartheta$. With $Q(p)=\gamma/(1-\gamma p)$, the value of
action $i$ when $p_i=p$, the regularized optimum at $x$ is $\pi^*(1|x)=\sigma(\vartheta u)$, where $\sigma$ is the logistic function
and $\beta u=Q(p_0+\Delta)-Q(p_0-\Delta)$. Since $1-\gamma p_0=\frac4{3H}$,
\begin{equation}\label{eq:gap}
\beta u=\frac{2\gamma^2\Delta}{(1-\gamma p_0)^2-\gamma^2\Delta^2}\ \ge\ \frac{2\gamma^2\Delta}{(1-\gamma p_0)^2}=\frac98\gamma^2H^2\Delta .
\end{equation}
The optimal value $V^*(x)=\beta\log(\frac12e^{Q(p_0+\Delta)/\beta}+\frac12e^{Q(p_0-\Delta)/\beta})$ does not depend on $\vartheta$, and
$V^*(x)\in[\frac23\gamma H,H]$, because $Q(p_0-\Delta)\ge\gamma/(\frac4{3H}+\frac1{6H})\ge\frac23\gamma H$. We write $\sigma(v)$ also for the
distribution on the two actions that puts mass $\sigma(v)$ on action $1$, and $\kl$ for the Bernoulli KL divergence.

\begin{lemma}[two-point loss]\label{lem:twopoint}
For $u>0$ and every distribution $q$ on the two actions, $\KL(q\|\sigma(u))+\KL(q\|\sigma(-u))\ge2\log\cosh(u/2)$; and
$\log\cosh v\ge0.3\min\{v^2,v\}$ for $v\ge0$.
\end{lemma}
\begin{proof}
The left side is convex in $q$ and symmetric under exchanging the actions, so it is minimized at the uniform $q$, where it
equals $-\log(4\sigma(u)\sigma(-u))=2\log\cosh(u/2)$. For $v\le1$, $\log\cosh v\ge v^2/2-v^4/12\ge\frac5{12}v^2$; for $v\ge1$,
$\log\cosh v\ge v-\log2\ge0.3v$, since $v-\log2-0.3v$ is increasing and positive at $v=1$.
\end{proof}

\begin{lemma}[information]\label{lem:info}
For $a,b\in\{p_0\pm\Delta\}$, $\kl(a,b)\le16\Delta^2/(1-p_0)=48\gamma H\Delta^2$.
\end{lemma}
\begin{proof}
$\kl(a,b)\le(a-b)^2/(b(1-b))$, $|a-b|\le2\Delta$, and $b(1-b)\ge\frac12\cdot\frac{1-p_0}2$ because $b\ge p_0-\Delta\ge\frac12$ and
$1-b\ge(1-p_0)/2$. Finally $1-p_0=(1-\gamma)/(3\gamma)$.
\end{proof}

\begin{lemma}[Assouad's lemma for gadgets]\label{lem:assouad}
Let $\{\mathcal M_\theta:\theta\in\{\pm1\}^J\}$ be MDPs, each containing $J$ gadgets with signs $\theta_1,\dots,\theta_J$ and a common $\Delta$, and
let $\Prob_\theta$ be the law of the data under $\mathcal M_\theta$. Suppose there are $\omega_j\ge0$, $\kappa_0\in(0,\infty]$ and, for every policy
$\pi$, distributions $q_j(\pi)$ on the two actions of gadget $j$ that do not depend on $\theta$, such that for all $\theta$ and $\pi$
\[
\SubOpt^\theta(\pi)\ \ge\ \sum_j\omega_j\,\beta\min\{\KL(q_j(\pi)\|\sigma(\theta_ju)),\kappa_0\}.
\]
If $\KL(\Prob_{\theta^{j+}}\|\Prob_{\theta^{j-}})\le48\gamma H\Delta^2\mathcal N$ for every $j$ and $\theta$, where $\theta^{j\pm}$ is $\theta$ with
coordinate $j$ set to $\pm1$, then every estimator $\hat\pi$ satisfies
\[
\max_\theta\E_\theta\,\SubOpt^\theta(\hat\pi)\ \ge\ \Big(\sum_j\omega_j\Big)\,\beta\min\{\log\cosh(u/2),\kappa_0/2\}\big(1-\sqrt{24\gamma H\Delta^2\mathcal N}\big).
\]
\end{lemma}
\begin{proof}
Write $\ell(q,\pm)=\beta\min\{\KL(q\|\sigma(\pm u)),\kappa_0\}$. Since $\min\{a,\kappa_0\}+\min\{b,\kappa_0\}\ge\min\{a+b,\kappa_0\}$,
Lemma~\ref{lem:twopoint} gives $\ell(q,+)+\ell(q,-)\ge\beta\min\{2\log\cosh(u/2),\kappa_0\}=:\ell_0$ for every $q$. Average over $\theta$
uniform on $\{\pm1\}^J$ and, for each $j$, pair $\theta^{j+}$ with $\theta^{j-}$:
$\E_{\theta^{j+}}\ell(q_j,+)+\E_{\theta^{j-}}\ell(q_j,-)\ge\int\min\{d\Prob_{\theta^{j+}},d\Prob_{\theta^{j-}}\}\,\ell_0=\ell_0(1-\TV_j)$, with
$\TV_j\le\sqrt{\KL/2}\le\sqrt{24\gamma H\Delta^2\mathcal N}$ by Pinsker's inequality. Summing over $j$ with weights $\omega_j$ bounds the average,
hence the maximum, from below by $\sum_j\omega_j\frac{\ell_0}2(1-\TV_j)$.
\end{proof}

\begin{lemma}[calibration]\label{lem:calib}
Let $\Omega=\sum_j\omega_j>0$. Suppose that for every $\Delta\in(0,\Delta_{\max}]$ a family $\{M_{\theta,\Delta}:\theta\in\{\pm1\}^J\}$ satisfies the
hypotheses of Lemma~\ref{lem:assouad} with the same $\mathcal N$, not depending on $\Delta$, and put
$L_1=\min\{0.00263\gamma^2H^2/\beta,\ 0.0281\gamma H\}$. If an estimator has $\max_{\theta,\Delta}\E_{\theta,\Delta}\SubOpt\le\epsilon$, the maximum over all
these families, with $\epsilon<\frac\Omega2\min\{L_1,\beta\kappa_0/2\}$, then
\[
\mathcal N\ \ge\ \min\Big\{4.94\cdot10^{-4}\,\frac{\Omega\gamma^3H^3}{\epsilon\beta},\ 7.41\cdot10^{-5}\,\frac{\Omega^2\gamma^3H^3}{\epsilon^2}\Big\}.
\]
\end{lemma}
\begin{proof}
If $\mathcal N\le N_0:=\frac38\gamma H$, take $\Delta=\Delta_{\max}$: then $24\gamma H\Delta^2\mathcal N\le\frac14$ and, by \eqref{eq:gap},
$u\ge u_m:=\frac{3\gamma H}{16\beta}$, so Lemma~\ref{lem:assouad} and $\log\cosh v\ge0.3\min\{v^2,v\}$ give
\[
\max_\theta\E\,\SubOpt\ \ge\ \tfrac\Omega2\min\{0.3\beta\min\{u_m^2/4,u_m/2\},\ \beta\kappa_0/2\}\ \ge\ \tfrac\Omega2\min\{L_1,\beta\kappa_0/2\}\ >\ \epsilon,
\]
a contradiction; here $0.3\beta u_m^2/4$ and $0.3\beta u_m/2$ are at least $0.00263\gamma^2H^2/\beta$ and $0.0281\gamma H$. So $\mathcal N>N_0$; take
$\Delta=(96\gamma H\mathcal N)^{-1/2}<\Delta_{\max}$, so that again $24\gamma H\Delta^2\mathcal N=\frac14$, and put
$u_0=\frac98\gamma^2H^2\Delta/\beta\le u$. Lemma~\ref{lem:assouad} gives
$\epsilon\ge\frac\Omega2\min\{0.3\beta\min\{u_0^2/4,u_0/2\},\beta\kappa_0/2\}$; the second entry exceeds $2\epsilon/\Omega$ by assumption, so
$0.15\,\Omega\beta\min\{u_0^2/4,u_0/2\}\le\epsilon$. If $u_0\le2$ this reads
\[
\frac{81}{64}\,\frac{\gamma^3H^3}{96\mathcal N\beta^2}\ \le\ \frac{\epsilon}{0.0375\,\Omega\beta},\quad\text{that is,}\quad \mathcal N\ \ge\ \frac{81\cdot0.0375}{64\cdot96}\,\frac{\Omega\gamma^3H^3}{\epsilon\beta};
\]
if $u_0>2$ it reads
\[
\frac98\,\frac{\gamma^{3/2}H^{3/2}}{\sqrt{96\mathcal N}}\ \le\ \frac{\epsilon}{0.075\,\Omega},\quad\text{that is,}\quad \mathcal N\ \ge\ \frac{(\frac98\cdot0.075)^2}{96}\,\frac{\Omega^2\gamma^3H^3}{\epsilon^2}.
\]
The two constants are $4.94\cdot10^{-4}$ and $7.41\cdot10^{-5}$ after rounding down.
\end{proof}

\begin{proof}[Proof of Theorem~\ref{thm:lb-design}]
Take $J$ gadgets $(x_j,y_{j1},y_{j2})$ with signs $\theta_j$ and a common $\Delta$, sharing $w$, and $\rho$ uniform on $\{x_j\}$. The pairs are
the $2J$ decision pairs, the $2J$ loop pairs and $w$, so $K=4J+1$.

\emph{Occupancies.} For every member, $d^*(x_j,i)=(1-\gamma)\pi^*(i|x_j)/J\le(1-\gamma)/J$. The killed chain reaches $y_{ji}$ with probability
at most $\gamma/J$ and then stops there with probability $(1-\gamma)/(1-\gamma p_i)\le(1-\gamma)\cdot\frac{6H}7=\frac67$, because
$1-\gamma p_i\ge1-\gamma p_0-\gamma\Delta_{\max}=\frac4{3H}-\frac1{6H}=\frac7{6H}$; so $d^*(y_{ji})\le\frac6{7J}$, and $d^*(w)\le1$.

\emph{Design.} $\mu(y_{ji})=\frac6{7JC}$, $\mu(x_j,i)=\frac{1-\gamma}{JC}$, and $\mu(w)$ the rest, which is at least
$1-\frac{12}{7C}-\frac{2(1-\gamma)}C\ge1-\frac37-\frac3{14}>\frac1C$ for $C\ge4$ and $\gamma\ge4/7$. Then $d^*\le C\mu$ at every pair for every member,
and $\mu$ does not depend on $(\theta,\Delta)$.

\emph{Loss.} Every killed path from $\rho$ visits exactly one decision state, at time $0$, and the other states have one action, so
by Lemma~\ref{lem:pathkl} and the chain rule, for every policy, $\SubOpt_\rho(\pi)=\frac1J\sum_j\beta\KL(\pi(\cdot|x_j)\|\sigma(\theta_ju))$. This is the
hypothesis of Lemma~\ref{lem:assouad} with $\omega_j=1/J$, $\kappa_0=\infty$ and $q_j(\pi)=\pi(\cdot|x_j)$.

\emph{Information.} The tuples are i.i.d.\ from a common design, so
$\KL(\Prob_{\theta^{j+}}\|\Prob_{\theta^{j-}})=n\sum_z\mu(z)\KL(P_{\theta^{j+}}(\cdot|z)\|P_{\theta^{j-}}(\cdot|z))$, and only $y_{j1},y_{j2}$ contribute: by
Lemma~\ref{lem:info} the sum is at most $n\cdot2\cdot\frac6{7JC}\cdot48\gamma H\Delta^2$. So Lemma~\ref{lem:assouad} holds, for every $\Delta$, with
$\mathcal N=\frac{12n}{7JC}$.

Lemma~\ref{lem:calib} with $\Omega=1$ gives $\mathcal N\ge7.41\cdot10^{-5}\gamma^3H^3/(\epsilon\max\{\epsilon,\beta\})$, that is,
$n\ge\frac{7JC}{12}\mathcal N$, which is the first bound. For the second, $\gamma^3\ge(4/7)^3$ and $J=(K-1)/4\ge K/5$.
\end{proof}

\section{Proofs for Section~\ref{sec:coverage}}\label{app:coverage}

All expectations in this appendix are over the killed chain of Appendix~\ref{app:killed}; $\E_*$ is under $\pi^*$ and $\E_{\mathrm{ref}}$
under $\pref$.

\begin{lemma}\label{lem:geom}
For $0<c<1$, $\E\,e^{cN/H}\le1/(1-c)$.
\end{lemma}
\begin{proof}
$\Prob(N=k)=(1-\gamma)\gamma^{k-1}$, so with $x=1/H$, $\E e^{cN/H}=xe^{cx}/(1-(1-x)e^{cx})$; the denominator is positive because
$\log(1-x)+cx\le-(1-c)x<0$. The claim is equivalent to $e^{y}(1-y)\le1$ for $y=cx\ge0$, which holds.
\end{proof}

\begin{lemma}[Bennett supermartingale]\label{lem:bennett}
If $0\le\nu\le1/H$, then $\mathcal E_t=\exp(\nu \Xi_t-(e-2)\nu^2V_t)$, with $\Xi_t=\sum_{k<t,\,k\le T}\xi_{k+1}$ and
$V_t=\sum_{k<t,\,k\le T}v_k$, is a nonnegative supermartingale for $(\mathcal F_t)$, and $\E\,\mathcal E_N\le1$.
\end{lemma}
\begin{proof}
For $x\le1$, $e^x\le1+x+(e-2)x^2$. Since $|\nu\xi|\le1$ and $\E[\xi_{t+1}|\mathcal G_t]=0$,
$\E[e^{\nu\xi_{t+1}}|\mathcal G_t]\le1+(e-2)\nu^2v_t\le e^{(e-2)\nu^2v_t}$, and $v_t$ is $\mathcal G_t$-measurable. Hence
$\E[\mathcal E_{t+1}|\mathcal F_t]\le\mathcal E_t$, conditioning first on $\mathcal G_t$. For the stopping time $N$, apply optional stopping to
$N\wedge m$ and let $m\to\infty$ with Fatou's lemma.
\end{proof}

\begin{lemma}[total variance along $\pi^*$]\label{lem:tvbound}
Let $v(z)=\gamma P(V^{*2})(z)-\gamma^2(PV^*(z))^2$, so that $v_t=v(z_t)$ in Lemma~\ref{lem:tele}. For every $\beta>0$ and every state $s$, along the
killed chain of $\pi^*$ started at $s$,
\[
\E_s\sum_{t\le T}\Big[\Var_{\pi^*(\cdot|s_t)}Q^*(s_t,\cdot)+v(z_t)\Big]\ \le\ 2H^2-V^*(s)^2 .
\]
Hence the expected remaining sum of $v$ from a pair, the pair included, is at most $2H^2$.
\end{lemma}
\begin{proof}
For each pair, $\gamma^2(PV^*)^2=(Q^*-r)^2\ge Q^{*2}-2rQ^*\ge Q^{*2}-2Hr$, so $v\le\gamma P(V^{*2})-Q^{*2}+2Hr$. Average over $a\sim\pi^*(\cdot|s)$, writing
$\pi^*f(s)=\sum_a\pi^*(a|s)f(s,a)$, and use $\pi^*(Q^{*2})=\Var_{\pi^*}Q^*+(\pi^*Q^*)^2$ and $\pi^*Q^*=V^*+\beta\KL(\pi^*\|\pref)\ge V^*\ge0$:
\[
\Var_{\pi^*}Q^*(s)+\pi^*v(s)\ \le\ \gamma P_{\pi^*}(V^{*2})(s)-V^*(s)^2+2H\,\pi^*r(s),
\]
with $P_{\pi^*}$ the state transition matrix of $\pi^*$. Along the killed chain $\E[V^*(s_{t+1})^2\mathbf 1\{t<T\}\,|\,s_t]=\gamma P_{\pi^*}(V^{*2})(s_t)$, so in
expectation the first two terms on the right telescope to $-V^*(s)^2$, and $\E_s\sum_{t\le T}\pi^*r(s_t)\le\E N=H$. From a pair $z$, the remaining sum is
at most $v(z)+\gamma P(2H^2-V^{*2})(z)=2\gamma H^2-\gamma^2(PV^*(z))^2\le2H^2$.
\end{proof}

\begin{lemma}[Khasminskii's lemma with pair conditioning]\label{lem:khas}
Let $V_N=\sum_{t\le T}v(z_t)$ under $\pi^*$. Then $\E_*V_N^j\le j!\,(2H^2)^j$, and $\E_*e^{sV_N}\le1/(1-2sH^2)$ for $0\le2sH^2<1$.
\end{lemma}
\begin{proof}
By Lemma~\ref{lem:tvbound}, conditionally on $\mathcal G_t$ the remaining sum including the current term has expectation at most $2H^2$. With
$x_t=v(z_t)\mathbf 1\{t\le T\}$, $V_N^j\le j!\sum_{t_1\le\dots\le t_j}x_{t_1}\cdots x_{t_j}$, and conditioning on $\mathcal G_{t_{j-1}}$ removes the last sum at the cost
$2H^2$; induction gives $\E_*V_N^j\le j!(2H^2)^j$, and summing the series gives the bound on the moment generating function.
\end{proof}

\begin{proposition}\label{prop:mgf}
For $c_2>\sqrt{32(e-2)}=4.7942\ldots$ and $\alpha=\beta/(c_2H)$,
\[
\log\E_*[L^\alpha]\ \le\ m(c_2):=\tfrac12\log\frac1{1-2/c_2}+\tfrac14\log\frac1{1-32(e-2)/c_2^2};
\]
in particular $\log\E_*[L^{\beta/(6H)}]\le m(6)=0.45708\ldots$.
\end{proposition}
\begin{proof}
Put $\lambda=\alpha/\beta=1/(c_2H)$ and $\lambda_2=2\lambda$. By Lemma~\ref{lem:tele} and $V^*\ge0$, $\alpha\log L\le\lambda(N+\Xi)$, so by the
Cauchy--Schwarz inequality $\E_*L^\alpha\le(\E_*e^{2\lambda N})^{1/2}(\E_*e^{\lambda_2\Xi})^{1/2}$. Lemma~\ref{lem:geom} with $c=2/c_2$ bounds the first
factor by $(1-2/c_2)^{-1/2}$. For the second, write $e^{\lambda_2\Xi}=(\mathcal E^{(2\lambda_2)}_N)^{1/2}\,e^{2(e-2)\lambda_2^2V_N}$, where $\mathcal E^{(2\lambda_2)}$
is the supermartingale of Lemma~\ref{lem:bennett} with $\nu=2\lambda_2=4/(c_2H)\le1/H$; the Cauchy--Schwarz inequality and
$\E_*\mathcal E^{(2\lambda_2)}_N\le1$ give $\E_*e^{\lambda_2\Xi}\le(\E_*e^{4(e-2)\lambda_2^2V_N})^{1/2}$, and Lemma~\ref{lem:khas} with $s=4(e-2)\lambda_2^2=16(e-2)/(c_2^2H^2)$, for
which $2sH^2=32(e-2)/c_2^2<1$, bounds the last expectation by $1/(1-32(e-2)/c_2^2)$. Collecting the exponents $\frac12$ and $\frac14$ gives $m(c_2)$. For
$c_2=6$ the two terms are $0.202733$ and $0.254354$, with sum $0.457087$.
\end{proof}

\begin{proof}[Proof of Theorem~\ref{thm:coverage}]
By Lemma~\ref{lem:lr} and H\"older's inequality with exponents $1+\alpha$ and $(1+\alpha)/\alpha$,
\[
\Prob^*(E)=\E_{\mathrm{ref}}[L\mathbf 1_E]\le(\E_{\mathrm{ref}}L^{1+\alpha})^{1/(1+\alpha)}\,\Prob_{\mathrm{ref}}(E)^{\alpha/(1+\alpha)},
\]
and $\E_{\mathrm{ref}}L^{1+\alpha}=\E_*L^\alpha$. Rearranging, $\Prob_{\mathrm{ref}}(E)\ge\Prob^*(E)^{1+1/\alpha}(\E_*L^\alpha)^{-1/\alpha}$. Take
$\alpha=\beta/(c_2H)$ and use Proposition~\ref{prop:mgf}: $(1/\alpha)\log\E_*L^\alpha\le c_2m(c_2)H/\beta=:c_1(c_2)H/\beta$, and $c_1(6)=2.7425\ldots\le2.75$.
For sets of pairs, take the event $\{z_T\in E\}$ and use Lemma~\ref{lem:occ}.
\end{proof}

\begin{remark}
The proof bounds the R\'enyi divergence of order $1+\beta/(c_2H)$ between the killed path laws by $c_1(c_2)H/\beta$; the inequality for events
alone would not give this moment bound. The route needs $c_2>4.79$; for example $c_1(5)=4.43$, $c_1(8)=2.05$, $c_1(10)=1.77$ and $c_1(50)=1.14$,
rounded up. It cannot reach $c_2\le2$, because the reward is bounded by the path length, whose exponential moment $\E e^{2N/(c_2H)}$ must be finite
uniformly in $H$.
\end{remark}

\begin{proof}[Proof of Corollary~\ref{cor:topj}]
Order the pairs of $E_\tau$ by $w=d^*/\dref$, $w_1\ge w_2\ge\cdots$, and let $F_j$ be the top $j$ of them. Then
$\dref(F_j)=\sum_{i\le j}d^*(z_i)/w_i\le d^*(F_j)/w_j$, and Theorem~\ref{thm:coverage} gives
$\dref(F_j)\ge e^{-c_1H/\beta}d^*(F_j)^{1+a}$, so $w_j\le e^{c_1H/\beta}d^*(F_j)^{-a}\le e^{c_1H/\beta}(j\tau)^{-a}$. Summing over $j$ gives the
first term; the second is Proposition~\ref{prop:cap}. For $a<1$, $\sum_{j\le m}j^{-a}\le m^{1-a}/(1-a)$.
\end{proof}

\begin{proof}[Proof of Proposition~\ref{prop:cap}]
Since $V^*(s)\in[\min_aQ^*(s,a),\max_aQ^*(s,a)]$ and $0\le Q^*\le H$, $|\log(\pi^*/\pref)|=|Q^*-V^*|/\beta\le H/\beta=:b$. The state
occupancy $d^\pi(s)=\sum_ad^\pi(s,a)$ satisfies $d^\pi=d^\pi B_\pi$ with $B_\pi=\gamma P_\pi+(1-\gamma)\mathbf 1\rho^\top$, a stochastic matrix that is
irreducible on the set $\cR$ of states reachable from $\rho$: from any state it jumps to the support of $\rho$ with probability at
least $1-\gamma$, and from there every state of $\cR$ is reachable; $\pi^*$ and $\pref$ have the same support, hence the same $\cR$. By
the Markov chain tree theorem \citep{pitman2016tree}, $d^\pi(s)=\omega_\pi(s)/\sum_{s'}\omega_\pi(s')$, where $\omega_\pi(s)$ is the sum, over spanning trees of $\cR$
directed to $s$, of the products of the $|\cR|-1$ edge weights. Each entry of $B_{\pi^*}$ is a positive combination of the corresponding
terms of $B_{\pref}$ with factors in $[e^{-b},e^b]$ (the reset term has factor $1$), so $\omega_{\pi^*}(s)/\omega_{\pref}(s)\in[e^{-(S_r-1)b},e^{(S_r-1)b}]$
and $d^*(s)/\dref(s)\le e^{2(S_r-1)b}$. Multiplying by the action ratio, at most $e^b$, gives the claim.
\end{proof}

\begin{proof}[Proof of Proposition~\ref{prop:det}]
With deterministic transitions and a fixed horizon there is no killing and no martingale term: the telescoping of
Lemma~\ref{lem:tele}, with stage-dependent values and $V^*_{H+1}=0$, gives $\beta\log L=R-V^*_1(s_1)\le R\le H$ for every trajectory.
\end{proof}

\section{Proofs for Section~\ref{sec:price} and Lemma~\ref{lem:chain}}\label{app:reference}

\begin{lemma}[information and counts for reference trajectories]\label{lem:traj}
Consider data of model (D2) with $n$ steps under a family of kernels $P_\theta$, and suppose that $\pref$, $\rho$ and the reset rule do
not depend on $\theta$. Let $n_z$ be the number of the first $n$ steps taken at the pair $z$. With known rewards,
(i) $\KL(\Prob_\theta\|\Prob_{\theta'})=\sum_z\E_\theta[n_z]\,\KL(P_\theta(\cdot|z)\|P_{\theta'}(\cdot|z))$, and
(ii) $\E_\theta[n_z]\le(n+H)\,d^{\pref}_{\rho,\theta}(z)$ for every $z$.
\end{lemma}
\begin{proof}
(i) The data are a sequence of transitions. Given the past, the next pair is $(s,a)$ with $s$ either the previous next state or,
after a reset, a draw from $\rho$, and $a\sim\pref(\cdot|s)$; only the law of the next state, $P_\theta(\cdot|z)$, depends on $\theta$. The
chain rule for the KL divergence over the $n$ steps gives the identity, as in the divergence decomposition for adaptively chosen
actions \citep{lattimore2020bandit}. There is no separate term for the pairs, which are generated by the recorded transitions.
(ii) The stream splits at resets into independent segments, each a killed path of $\pref$ from $\rho$, whose expected number of
steps at $z$ is $\sum_t\gamma^t\Prob_{\mathrm{unkilled}}(z_t=z)=H\,d^{\pref}_{\rho,\theta}(z)$. The number of segments that start among the first
$n$ steps is a stopping time for the segment sequence with mean at most $1+n(1-\gamma)$, since every step is followed by a reset
with probability $1-\gamma$. Wald's identity gives $\E_\theta[n_z]\le(1+n/H)\,H\,d^{\pref}_{\rho,\theta}(z)$.
\end{proof}

\paragraph{The chain.} The chain of Lemma~\ref{lem:chain} has states $c_0,\dots,c_{m-1}$ with zero rewards and starts at $c_0$. At $c_k$
the action go has reference probability $\frac12$ and moves, with probability $\frac12$, to $c_{k+1}$ (from $c_{m-1}$, to an end state
whose value lies in $[0,H]$) and otherwise to an absorbing state of reward $1$, whose value is $H$; the action quit has reference
probability $\frac12$ and moves to an absorbing zero-reward state. Let $g_k=\pi^*(\mathrm{go}|c_k)$.

\begin{proof}[Proof of Lemma~\ref{lem:chain}]
Let $Q_k=Q^*(c_k,\mathrm{go})=\gamma(\frac12V_{k+1}+\frac12H)$, where $V_{k+1}\in[0,H]$ is the value of the next state; so
$Q_k\in[\gamma H/2,\gamma H]$ and $Q_k\le\beta$. Quit has $Q^*=0$, so $V_k=\beta\log(\frac12e^{Q_k/\beta}+\frac12)$, and
$\log((1+e^x)/2)=x/2+\log\cosh(x/2)\in[x/2,\,x/2+x^2/8]$ gives $Q_k/2\le V_k\le Q_k/2+Q_k^2/(8\beta)$. Hence
$\log(2g_k)=(Q_k-V_k)/\beta\ge(Q_k/2-Q_k^2/(8\beta))/\beta\ge3Q_k/(8\beta)\ge\frac{3\gamma}{16}\frac H\beta$, and
$g_k\le\frac12e^{Q_k/(2\beta)}\le\frac12e^{1/2}$, so the factor $\gamma g_k/2$ by which the reach probability of $\pi^*$ shrinks at $c_k$
lies in $[\gamma/4,\,\gamma e^{1/2}/4]\subseteq[\gamma/4,0.42]$. So $\delta_*=\prod_k\gamma g_k/2\in[(\gamma/4)^m,0.42^m]$ and
$\delta_{\mathrm{ref}}=(\gamma/4)^m$. Then $\log(\delta_*/\delta_{\mathrm{ref}})=\sum_k\log(2g_k)\ge m\,\frac{3\gamma}{16}\frac H\beta$ and
$\log(1/\delta_*)\le m\log(4/\gamma)$; combining the two gives $\log(\delta_*/\delta_{\mathrm{ref}})\ge\kappa\frac H\beta\log(1/\delta_*)$.
\end{proof}

\begin{proof}[Proof of Theorem~\ref{thm:power}]
We prove the more precise bound: for $J\ge1$ and $K=2m+4J+4$, with $m\le\bar m$ chosen below, every estimator with
$\max_\theta\E_\theta[\SubOpt_\rho]\le\epsilon$ needs
\begin{equation}\label{eq:power-precise}
n\ \ge\ \frac7{16\gamma}\,HJ\Big(\frac{\gamma^4H^2}{6072\,\epsilon\beta}\Big)^{1+\kappa H/\beta}-H .
\end{equation}

\emph{Construction.} The chain above, followed by an end state $e$ with one action and reward $0$, which moves to $x_j$ with probability
$1/J$ for each of $J$ gadgets with signs $\theta_j$ and $\Delta=\Delta_{\max}$, sharing the sink $w$; $\rho=\delta_{c_0}$. The pairs are $2m$ on the
chain, $e$, the two absorbing states of the chain, $w$, and $4J$ in the gadgets. Since $V^*(x_j)$ does not depend on $\theta_j$, neither do
$V^*(e)\in[0,H]$, the optimum on the chain, nor the laws of the chain under $\pi^*$ and $\pref$. Let $\delta_*(m)$ and $\delta_{\mathrm{ref}}(m)$
be the probabilities that the killed chains of $\pi^*$ and $\pref$ reach $e$. Index the chain states by their distance $d=1,\dots,m$ to
$e$: the go probability $g^{(d)}$ at distance $d$ depends only on $d$, because values are computed backward from $e$, so
$\delta_*(m)=\prod_{d\le m}\gamma g^{(d)}/2$ and $\delta_*(m+1)=\delta_*(m)\,\gamma g^{(m+1)}/2$. By the proof of Lemma~\ref{lem:chain} every factor
lies in $[\gamma/4,0.42]$ and $\delta_{\mathrm{ref}}\le\delta_*^{1+\kappa H/\beta}$. Choose $m$ as the largest integer with $\delta_*(m)>\delta_\epsilon$. It is
at least $1$ because $\delta_*(1)\ge\gamma/4>\delta_\epsilon$, it is at most $\bar m$ because $\delta_\epsilon<\delta_*(m)\le0.42^m$, and
$\delta_*(m)\le\frac4\gamma\delta_*(m+1)\le\frac4\gamma\delta_\epsilon$.

\emph{Loss.} Let $\cR$ be the event that the path reaches $e$, and for a policy $\pi$ let $q_j(\pi)$ be the average of the action
distribution at $x_j$ over the histories that reach $x_j$, weighted by their probabilities under $\pi$ and the true kernel, which on
the chain does not depend on $\theta$. By Lemma~\ref{lem:pathkl}, $\SubOpt^\theta(\pi)=\beta\KL(\Prob^\pi\|\Prob^*)$. If $\Prob^\pi(\cR)\le\delta_*/2$, data
processing and $\kl(a,\delta)\ge\kl(\delta/2,\delta)\ge\frac{1-\log2}2\delta$ for $a\le\delta/2$ give $\SubOpt\ge0.153\beta\delta_*$. Otherwise each $x_j$ is
reached with probability $\gamma\Prob^\pi(\cR)/J>\gamma\delta_*/(2J)$, and the chain rule for the path KL, the nonnegativity of its terms, and
convexity of the KL divergence in its first argument give $\SubOpt\ge\sum_j\frac{\gamma\delta_*}{2J}\beta\KL(q_j(\pi)\|\sigma(\theta_ju))$. In both
cases $\SubOpt\ge\sum_j\omega_j\beta\min\{\KL(q_j(\pi)\|\sigma(\theta_ju)),\kappa_0\}$ with $\omega_j=\frac{\gamma\delta_*}{2J}$ and $\kappa_0=\frac{0.306}\gamma$, because
$\sum_j\omega_j\beta\kappa_0=0.153\beta\delta_*$.

\emph{Information.} By Lemmas~\ref{lem:traj} and~\ref{lem:info}, $\KL(\Prob_{\theta^{j+}}\|\Prob_{\theta^{j-}})\le48\gamma H\Delta^2\mathcal N$ with
$\mathcal N=\max_\theta\E_\theta[n_{y_{j1}}+n_{y_{j2}}]\le(n+H)\max_\theta\sum_i\dref(y_{ji})$. The reference chain reaches $y_{ji}$ with probability
$\delta_{\mathrm{ref}}\cdot\frac\gamma J\cdot\frac12\cdot\gamma$ and then stops there with probability at most $\frac67$, so
$\sum_i\dref(y_{ji})\le\frac{6\gamma^2}7\frac{\delta_{\mathrm{ref}}}J$ and $\mathcal N\le\frac{6\gamma^2}{7J}(n+H)\delta_{\mathrm{ref}}$.

\emph{Conclusion.} Suppose $\mathcal N\le N_0=\frac38\gamma H$. Then $24\gamma H\Delta_{\max}^2\mathcal N\le\frac14$, and since $\beta\ge H$,
$u\ge u_m=\frac{3\gamma H}{16\beta}$ with $u_m\le\frac3{16}$, so $\log\cosh(u/2)\ge0.075u_m^2=0.075\cdot\frac9{256}\gamma^2H^2/\beta^2<\kappa_0/2$.
Lemma~\ref{lem:assouad} gives $\max_\theta\E\,\SubOpt\ge a\gamma^3\delta_*H^2/\beta$ with $a=\frac12\cdot\frac12\cdot0.075\cdot\frac9{256}=\frac{27}{40960}\approx6.6\cdot10^{-4}$, and
$a\gamma^3\delta_*H^2/\beta>\epsilon$ because $\delta_*>\delta_\epsilon$ and $1518\,a>1$. So an estimator with error at most $\epsilon$ has $\mathcal N>N_0$,
that is, $(n+H)\delta_{\mathrm{ref}}>\frac{7J}{6\gamma^2}\cdot\frac38\gamma H=\frac{7HJ}{16\gamma}$. With $\delta_{\mathrm{ref}}\le\delta_*^{1+\kappa H/\beta}$ and
$\delta_*\le\frac4\gamma\delta_\epsilon=\frac{6072\,\epsilon\beta}{\gamma^4H^2}$ this is \eqref{eq:power-precise}.

\emph{The form in the theorem.} Take $J=\lfloor(K-2m-4)/4\rfloor$ and add at most three isolated zero-reward states. Since
$g^{(d)}=\sigma(Q^{(d)}/\beta)\le\sigma(1)<\frac34$, $\delta_*(m)<(3/8)^m$, so $\delta_\epsilon<(3/8)^m$ and $\bar m>1+1.1306\,m$. For $m\ge2$ the condition
$K\ge2(2\bar m+4)$ then gives $K\ge4m+14$, and for $m=1$ (where $\delta_\epsilon<\frac14$) it gives $K\ge19$. In both cases $J\ge(K-2m-7)/4\ge K/8$.
Write the power in \eqref{eq:power-precise} as $b\cdot b^{\kappa H/\beta}$ with $b=\gamma^4H^2/(6072\epsilon\beta)$; then the first
term of \eqref{eq:power-precise} is $\frac{7\gamma^3J}{16\cdot6072}\,\frac{H^3}{\epsilon\beta}\,b^{\kappa H/\beta}$. With $J\ge K/8$ and $\gamma\ge4/7$,
$\frac{7\gamma^3J}{16\cdot6072}\ge\frac{7(4/7)^3}{128\cdot6072}K>1.68\cdot10^{-6}K$, and $b\ge c'H^2/(\epsilon\beta)$ with $c'=(4/7)^4/6072>1.75\cdot10^{-5}$.
\end{proof}

\begin{proof}[Proof of Theorem~\ref{thm:gate}]
\emph{Construction.} Put $C'=C/2\ge2$. Take $J$ gadgets $(x_j,y_{j1},y_{j2})$ with signs $\theta_j$ and a common $\Delta$, sharing the sink $w$,
and in front of gadget $j$ a gate state $g_j$ with two actions: enter, with $\pref(\mathrm{enter}|g_j)=1/C'$, moves to $x_j$; skip, with
$\pref=1-1/C'$, moves to an absorbing zero-reward state $o$ shared by all gates. Rewards at $g_j$ are $0$, $\rho$ is uniform on $\{g_j\}$,
and $\pref$ is uniform at the decision states. The pairs are $2J$ at the gates, $2J$ at the decision states, $2J$ loop pairs, $o$ and
$w$, so $K=6J+2$. Only the loop transitions depend on $(\theta,\Delta)$.

\emph{The gate.} $Q^*(g_j,\mathrm{enter})=\gamma V^*(x_j)\ge\frac23\gamma^2H$ and $Q^*(g_j,\mathrm{skip})=0$, so $e_*:=\pi^*(\mathrm{enter}|g_j)=\sigma(\ell)$ with
$\ell=\gamma V^*(x_j)/\beta-\log(C'-1)\ge\gamma^2H/(3\beta)>0$ by the condition on $C$. Hence $e_*\ge\frac12$, and $e_*$ does not depend on $\theta$.

\emph{Coverage.} Every path visits one gate, at time $0$, with probability $1/J$, and a gadget only after entering. The ratios of
occupancies are: at (gate, enter) $e_*C'\le C'$; at (gate, skip) and at $o$, $(1-e_*)/(1-1/C')\le1$, since $e_*\ge\frac12$ and $C'\ge2$; at
$(x_j,i)$ and at the loop $y_{ji}$, $e_*\pi^*(i|x_j)/(\frac1{C'}\cdot\frac12)\le2C'$, because the loop kernel is the same under both policies;
at $w$, a mixture of the loop terms with the same weights, so at most $2C'$. Hence $C^*(\pref)\le2C'=C$ for every member, and the enter
ratio $e_*C'\ge C/4$ gives $C^*(\pref)\ge C/4$.

\emph{Loss.} By Lemma~\ref{lem:pathkl}, $\SubOpt_\rho(\pi)=\beta\KL(\Prob^\pi\|\Prob^*)$ for every policy. The only states with two actions on a
path are its gate and, after enter, its decision state, which is reached by one history. Writing $a_j=\pi(\mathrm{enter}|g_j)$ and $q_j(\pi)$
for the action distribution at $x_j$ after $(g_j,\mathrm{enter})$, the chain rule gives
\[
\SubOpt_\rho(\pi)=\frac\beta J\sum_j\Big[\kl(a_j,e_*)+\gamma a_j\KL\big(q_j(\pi)\|\sigma(\theta_ju)\big)\Big].
\]
If $a_j\le\frac14$, then $\kl(a_j,e_*)\ge\kl(\frac14,e_*)$, because $\kl(\cdot,e_*)$ decreases on $[0,e_*]$ and $\frac14<\frac12\le e_*$. With
$\mathsf h=-\frac14\log\frac14-\frac34\log\frac34=0.5623\ldots$,
\[
\kl(\tfrac14,\sigma(\ell))=-\mathsf h+\tfrac14\log(1+e^{-\ell})+\tfrac34\log(1+e^{\ell})\ \ge\ \tfrac34\ell-\mathsf h ,
\]
and $\kl(\frac14,e_*)\ge\kl(\frac14,\frac12)=0.1308\ldots$ since $e_*\ge\frac12$. So $K_g:=\beta\kl(\frac14,e_*)\ge\max\{0.1308\beta,\ \frac{\gamma^2H}4-0.5624\beta\}$.
The two entries are equal at $\beta=\gamma^2H/(4\cdot0.6932)$, which gives $K_g\ge0.0471\,\gamma^2H$ for every $\beta$. If $a_j>\frac14$, the second term
is at least $\frac\gamma4\beta\KL(q_j(\pi)\|\sigma(\theta_ju))$. In both cases the $j$-th term is at least
$\frac{\gamma\beta}4\min\{\KL(q_j(\pi)\|\sigma(\theta_ju)),\kappa_0\}$ with $\kappa_0:=0.1884\,\gamma H/\beta\le4K_g/(\gamma\beta)$, which does not depend on
$(\theta,\Delta)$. This is the hypothesis of Lemma~\ref{lem:assouad} with $\omega_j=\gamma/(4J)$ and $\Omega=\gamma/4$.

\emph{Information.} By Lemma~\ref{lem:traj}(i) only the loop pairs of gadget $j$ contribute to $\KL(\Prob_{\theta^{j+}}\|\Prob_{\theta^{j-}})$, which by
Lemma~\ref{lem:info} is at most $48\gamma H\Delta^2\,\E_\theta[n_{y_{j1}}+n_{y_{j2}}]$. The reference chain is at $x_j$ at time $1$ with probability
$\gamma/(JC')$, at $y_{ji}$ at time $2$ with probability $\gamma^2/(2JC')$, and then spends expected discounted time at most
$1/(1-\gamma p_i)\le\frac{6H}7$ there, so $\sum_i\dref(y_{ji})\le\frac{6\gamma^2}{7JC'}$, and by Lemma~\ref{lem:traj}(ii) the hypothesis of
Lemma~\ref{lem:assouad} holds, for every $\Delta$, with $\mathcal N=\frac{6\gamma^2}{7JC'}(n+H)=\frac{12\gamma^2}{7JC}(n+H)$.

\emph{Conclusion.} Lemma~\ref{lem:calib} applies with $\Omega=\gamma/4$: its condition $\epsilon<\frac\Omega2\min\{L_1,\beta\kappa_0/2\}$ is the stated
one, because $\beta\kappa_0/2=0.0942\,\gamma H>0.0281\,\gamma H\ge L_1$. It gives
$\mathcal N\ge\min\{4.94\cdot10^{-4}\,\frac\gamma4\,\frac{\gamma^3H^3}{\epsilon\beta},\ 7.41\cdot10^{-5}\,\frac{\gamma^2}{16}\,\frac{\gamma^3H^3}{\epsilon^2}\}$, and multiplying by
$7JC/(12\gamma^2)$ gives the first display of the theorem (the coefficients are $7.210\cdot10^{-5}$ and $2.703\cdot10^{-6}$). For the second,
$\gamma\ge4/7$, $\min\{a/(\epsilon\beta),b/\epsilon^2\}\ge\min\{a,b\}/(\epsilon\max\{\epsilon,\beta\})$ and $J=(K-2)/6\ge K/8$; the constant is
$2.703\cdot10^{-6}(4/7)^3/8=6.30\ldots\cdot10^{-8}\ge6.3\cdot10^{-8}$.
\end{proof}

\section{Proof of Theorem~\ref{thm:plugin}}\label{app:plugin}

On $I(m,\gamma)$, $K=m+3$, $Q^*(x,a_0)=\frac34\gamma H$ for the main action $a_0$, $Q^*(x,a_i)=\frac12\gamma H$ for the rare actions
$a_1,\dots,a_m$, and the gap is $\Delta=\gamma H/4$. Rewards are deterministic, so $\hat r=r$ at every seen pair, and the plug-in policy is
the soft-optimal policy of $(\hat P,r)$, with $\hat Q=0$ on unseen pairs.

\begin{proof}
\emph{Coverage.} $d^*(x,a)/\dref(x,a)=\pi^*(a|x)/\pref(a|x)$. With $q=\pi^*(\{a_1,\dots,a_m\}|x)=1/(1+e^{\Delta/\beta})$, a direct computation gives
$C_\Sigma(\cZ)=\frac{58}{15}+(2m-\frac{26}{15})q$. Since $\beta\le\Delta/\log m$, $q\le e^{-\Delta/\beta}\le1/m$ and $C_\Sigma(\cZ)\le\frac{58}{15}+2<6$.

\emph{Loss.} The only state with a choice is $x$, visited at time $0$, so Lemma~\ref{lem:pathkl} gives
$\SubOpt_\rho(\pi)=\beta\KL(\pi(\cdot|x)\|\pi^*(\cdot|x))$. For a set $\mathcal X$ of rare actions, data processing to the partition $\{\mathcal X,\mathcal X^c\}$ gives
$\SubOpt\ge\beta\,\kl(\pi(\mathcal X),\pi^*(\mathcal X))$. Here $\pi^*(a_i|x)=\frac1{2m}e^{\gamma H/(2\beta)}/Z^*\le\frac1me^{-\gamma H/(4\beta)}$, because
$Z^*\ge\frac12e^{3\gamma H/(4\beta)}$; so $\pi^*(\mathcal X)\le e^{-\gamma H/(4\beta)}$. If $\pi(\mathcal X)\ge\frac12$, then
$\kl(\pi(\mathcal X),\pi^*(\mathcal X))\ge\kl(\frac12,\pi^*(\mathcal X))\ge\frac12\log(1/\pi^*(\mathcal X))-\log2$, and
\[
\SubOpt\ \ge\ \frac{\gamma H}8-\beta\log2\ \ge\ \frac{\gamma H}8\Big(1-\frac{\log2}{\log80}\Big)\ \ge\ 0.105\,\gamma H,
\]
using $\beta\le\gamma H/(8\log80)$, which holds because $k_0\ge8$.

\emph{Counts.} Let $N_x$ be the number of recorded transitions from $x$. Under (D2$'$), $N_x\sim\mathrm{Bin}(n,1/H)$; under (D2),
$N_x=1+\mathrm{Bin}(n-1,1/H)$, because a segment starts at step $0$ and after each step with probability $1-\gamma$, and the first transition of
each segment is from $x$. In both models, given $N_x$, the actions of these transitions are i.i.d.\ from $\pref(\cdot|x)$ and their next
states are independent draws from $P(\cdot|x,a)$. By Chernoff's inequality \citep{boucheron2013concentration}, $n/(2H)\le N_x\le2n/H$ except
with probability $2e^{-n/(10H)}$ (for (D2) use $2n/H-1\ge1.9(n-1)/H$, which holds for $n\ge2000H$). Condition on such a value. Each rare
action's count is $\mathrm{Bin}(N_x,q_0)$ with $q_0=1/(2m)\le1/4000$ and mean $\lambda=N_x/(2m)\in[0.2498,1]$, since $n/H\le m\le n/H+1$. Let $U$ be the
number of rare actions with exactly one sample and $\mathcal X'$ those among them whose sample moved to $G$. Then
$\E[U|N_x]=m\lambda(1-q_0)^{N_x-1}\ge m\lambda e^{-\lambda/(1-q_0)}\ge0.194\,m$. Changing one of the $N_x$ i.i.d.\ actions changes $U$ by at most $2$, so
McDiarmid's inequality gives $\Prob(U<0.1m\,|\,N_x)\le\exp(-(0.094m)^2/(2N_x))\le e^{-0.0022m}\le e^{-4.4}$, using $N_x\le2m$ and $m\ge2000$. Given the
counts, the outcomes of the single samples are independent fair coins, so $\Prob(|\mathcal X'|<U/4\,|\,U)\le e^{-U/8}$ and
$\Prob(|\mathcal X'|<m/40\,|\,N_x)\le e^{-4.4}+e^{-m/80}$. By the choice of $k_0$ and $\Prob(\mathrm{Bin}(N_x,q_0)\ge k)\le(e\lambda/k)^k\le(e/k)^k$, every rare action has
at most $k_0$ samples except with probability $0.05$. The main action's count is $\mathrm{Bin}(N_x,\frac12)$; by Chernoff's and Hoeffding's
inequalities it is at least $N_x/4$ and its empirical probability of $G$ is at most $\frac78$, except with probability $e^{-n/(32H)}+e^{-n/(256H)}$.
Finally both absorbing pairs are seen except with probability $2e^{3/16}e^{-3n/(32H)}$. Under (D2$'$) this holds because $\dref(G)=\frac58\gamma$
and $\dref(B)=\frac38\gamma$. Under (D2), on $\{N_x\ge n/(2H)\}$ the first $N_x-1$ segments are complete, so the event that $B$ is unseen and
$N_x\ge n/(2H)$ is contained in the event that none of the first $\lceil n/(2H)\rceil-1$ segments of the i.i.d.\ segment sequence records a
transition from $B$; each does with probability at least $\frac38\gamma$ (at least $\frac58\gamma$ for $G$), which gives the bound. With $n\ge2000H$ the
total exceptional probability is at most $0.0123+0.05+e^{-7.8}+e^{-25}<0.07$.

\emph{On the good event.} $\hat V(G)=H$ and $\hat V(B)=0$. Let $\mathcal X$ be the set of seen rare actions whose samples all moved to $G$; then
$\mathcal X\supseteq \mathcal X'$, $|\mathcal X|\ge m/40$, and $\hat Q(x,a)=\gamma H$ on $\mathcal X$. Every other rare action has $\hat Q(x,a)\le\gamma H(1-1/k_0)$ (a seen one has at most
$k_0$ samples and at least one moved to $B$; an unseen one has $\hat Q=0$), these actions have total reference mass at most $\frac12$, and
$\hat Q(x,a_0)\le\frac78\gamma H$. So the plug-in policy satisfies
\[
\frac{\pi_{\mathrm{PI}}(\mathcal X)}{1-\pi_{\mathrm{PI}}(\mathcal X)}\ \ge\ \frac{\frac{|\mathcal X|}{2m}e^{\gamma H/\beta}}{\frac12e^{7\gamma H/(8\beta)}+\frac12e^{\gamma H(1-1/k_0)/\beta}}
\ \ge\ \frac{1/40}{e^{-\gamma H/(8\beta)}+e^{-\gamma H/(k_0\beta)}}\ \ge\ 1,
\]
since each exponential is at most $1/80$ by the condition on $\beta$. Hence $\pi_{\mathrm{PI}}(\mathcal X)\ge\frac12$, and $\SubOpt\ge0.105\gamma H$ on an event of
probability at least $0.93$, so $\E\,\SubOpt\ge0.097\,\gamma H$.
\end{proof}

\begin{remark}[large $\gamma H/\beta$]\label{rem:plugin}
The proof uses the condition $\beta\le\gamma H/(k_0\log80)$ only through $e^{-B/8}+e^{-B/k_0}\le1/40$ in the odds and $B\ge8\log80$ in the loss,
where $B=\gamma H/\beta$. For $B\ge40$ both hold whenever $\beta\le\gamma H/(4\log m)$. The second is immediate. For the first: if $k_0\le9$ the sum is
below $0.019$; if $k_0=10$, minimality of $k_0$ gives $\log m>\log0.05+10(\log10-1)$, so $B\ge4\log m>40.12$ and the sum is below $0.0248$; if
$k_0\ge11$, minimality gives $B/k_0>4(\log k_0-1)+4\log(0.05)/k_0\ge4.50>\log80$. So for $\gamma H/\beta\ge40$ the plug-in fails for every
$n\ge2000H$ with $\lceil n/H\rceil\le e^{\gamma H/(4\beta)}$.
\end{remark}

\paragraph{PSVI on $I(m,\gamma)$.} Under the condition on $\beta$ of Theorem~\ref{thm:plugin}, the set $E=\{(x,a_0),G,B\}$, with $a_0$ the main action, has
$C_\Sigma(E)\le58/15$ and $d^*(E^c)\le1/n$. Theorem~\ref{thm:h4}(a) then bounds the loss of PSVI under (D2$'$) by
$\cu(\min\{H^2\sqrt{\iota/n},\,H^4\iota\Lambda/(\beta n)\}+H^2/n)$ with probability at least $1-2\delta$ once $n\ge32H\iota$, and Theorem~\ref{thm:h4}(b) gives the
same for the subsampled estimator under (D2), with $n'$ in place of $n$ in the first term, once $n\ge10^5H^2\iota$. At fixed $H$ and with $\delta=1/n$,
these bounds tend to $0$ as $n$ grows.

\section{Numerical computations}\label{app:checks}

We evaluated the main inequalities numerically on small instances. Three kinds of computation appear below: exact evaluations of the
constructions of the lower bounds and of the coverage bounds; evaluations of the inequalities of Appendices~\ref{app:killed}
and~\ref{app:upper} on small random MDPs, where the linear systems of the killed chain give every expectation exactly; and simulations of the
estimators. Table~\ref{tab:checks} lists what we measured, usually the largest ratio of the left side of an inequality to its right side.
``Random MDPs'' have Dirichlet transition kernels and uniform mean rewards. All of the computations together took about
\TotalCountedMinutes\ minutes on a 4-core CPU.

\begin{table}[p]
\caption{Numerical computations for the paper's statements. Unless a row says otherwise, the value is the worst case over all instances, and a ratio is that of the left side of an inequality to its right side. For the loop family the table gives least-squares
slopes of the mean loss in $\log H$, with the range over three repetitions where they differ; for Theorem~\ref{thm:plugin} it gives the smallest mean loss over the sample sizes. The
paragraphs of this appendix describe the instances.}
\label{tab:checks}
\centering
\footnotesize
\begin{tabular}{@{}>{\raggedright\arraybackslash}p{2.4cm}>{\raggedright\arraybackslash}p{3.9cm}>{\raggedright\arraybackslash}p{6.7cm}@{}}
\toprule
Statement & Instances & Measured \\
\midrule
Lemmas~\ref{lem:pathkl}, \ref{lem:pathpess} & random MDPs and slacks & ratio of $\SubOpt$ to $\beta\E_*\varphi(2Z/\beta)$ at most $0.43$; path-KL identity to $2.4\cdot10^{-8}$ \\
Lemma~\ref{lem:contraction} & 60000 draws, variance term active in 43\%, $\omega\in[0.01,0.6]$ & smallest partial derivative $0$ (negative in 3003 draws with the floor halved); no decrease along $V\le V'$; no Lipschitz excess beyond rounding; additive penalty: derivative sums up to $1.61$ where unclipped \\
Proposition~\ref{prop:eventB} & random MDPs, 27 settings of 100 datasets each, $\delta=0.1$ & deviation at most $0.14$ of the penalty; $\hat\sigma^2$ at most $0.37$ of $2\sigma^2+18H^2\iota/n_z$ \\
Proposition~\ref{prop:eventB} & self-loop instance, $H\in\{2,4,8\}$, $n\in\{2\cdot10^5,\dots,2\cdot10^7\}$, 27 settings of 100 datasets each & variance term active at both pairs of $s_0$ in every dataset; deviation at most $0.19$ of the penalty; $\hat\sigma-\sigma$ at most $0.04$ of $3H\sqrt{\iota/n_z}$ \\
Proposition~\ref{prop:subsample} & Hoeffding penalty, trajectory streams, $H\in\{2,4\}$, 240 datasets & deviation at most $0.15$ of the penalty \\
Propositions~\ref{prop:eventB}, \ref{prop:subsample} & Bernstein penalty, self-loop instance under (D2), $H\in\{2,4\}$, 240 datasets & deviation at most $0.15$ of the penalty; $\hat\sigma-\sigma$ at most $0.03$ of $3H\sqrt{\iota/n_z}$ \\
proof of Theorem~\ref{thm:h3} (energy step, cross term) & 900 random MDPs, $\beta/H\in\{8,16,100\}$, pessimistic $\hat Q$ from random penalties & energy step at most $0.5$ of its bound, cross term at most $0.47$ \\
Lemmas~\ref{lem:loctilt}--\ref{lem:avp}, Proposition~\ref{prop:eventB}(ii) & 1200 random MDPs, eight values of $\beta/H$ from $1.32$ to $16$; 1080 sampled datasets & tilt bounds at most $0.99$ of their constants; energy inequalities at most $0.1$; $\E_s\Gamma\le\bar\Gamma$ and the bounds for $\hat V$ at most $0.03$; Lemma~\ref{lem:ptv} at most $0.62$; Lemma~\ref{lem:avp} at most $0.73$; in Proposition~\ref{prop:eventB}(ii), $\hat\sigma-\sigma$ at most $0.63$ of $3H\sqrt{\iota/n_z}$ \\
Corollary~\ref{cor:refdata} & random MDPs and chains of length up to $40$, $\beta/H\in\{0.5,1,1.4,2,8\}$, $n$ from $10^3$ to $10^{24}$, exact occupancies & count condition at most $0.56$ and coverage sum at most $0.21$ of their bounds (the other two steps are arithmetic); without the power (same $c_1$) the count condition fails on long chains \\
Theorem~\ref{thm:h3} (energy step), Lemma~\ref{lem:mixed} & 900 pessimistic fixed points, $\beta/H$ from $0.1$ to $16$ & Young step at most $0.15$ ($\lambda=1$) and $0.06$ ($\lambda=9$); Lemma~\ref{lem:mixed} at most $0.02$ \\
Proposition~\ref{prop:burnin} (split) & $\beta/H\in\{0.2,1\}$, $n\in\{200,2000,20000\}$, 1080 sampled datasets & slack at most $0.09$ of the split with $c_s=72$ at every seen pair \\
Theorems~\ref{thm:h4}--\ref{thm:h3} & loop family, $\beta=8H$, $H\in\{8,\dots,64\}$, 120 datasets per $H$ & $H$-slope of the mean loss: Bernstein $2.16$ (local slopes 2.33, 2.14, 2.02), Hoeffding $3.06$, plug-in $1.8$--$2$ \\
Theorems~\ref{thm:lb-design},~\ref{thm:power} & exact, over grids of $\gamma,\beta,\Delta$ & largest ratio in the inequalities of the constructions: $1-9.4\cdot10^{-9}$ \\
Theorem~\ref{thm:gate} & exact, over grids of $\gamma,\beta,C,\Delta$ & largest ratio in the inequalities of the construction: $1-9.4\cdot10^{-9}$ \\
Lemma~\ref{lem:chain} & chains, $\beta\ge H$ & chain exponent over $\kappa H/\beta\ge1.73$ \\
Theorem~\ref{thm:coverage}, Proposition~\ref{prop:mgf} & 900 random MDPs, 2--8 states, $\beta/H\in\{0.02,0.1,1,10\}$ & $\log\E_*L^{\beta/(6H)}$ at most $0.077$ of the bound $m(6)$ \\
Appendix~\ref{app:constants} & loop family at $n=4\cdot10^6$; instance of Theorem~\ref{thm:plugin} at $n\le2\cdot10^5$ & explicit bound of Theorem~\ref{thm:h3} at least $3.4\cdot10^{4}$ times the reference policy's loss; of Theorem~\ref{thm:h4} at least $11.5$ times it \\
Theorem~\ref{thm:plugin} & $H=20$, $\beta=0.4$, $n\in\{2\cdot10^3,2\cdot10^4,2\cdot10^5\}$ ($n\ge2000H$ only at $2\cdot10^5$), 600 datasets per $n$ & smallest mean plug-in loss $0.27H$, under (D2$'$) and (D2) \\
\bottomrule
\end{tabular}

\end{table}

\paragraph{The loop family (Section~\ref{sec:experiments}).} A decision state $x$ has a uniform reference policy. Action $A$ moves to a loop state $y$
(reward $1$, stay probability $1-2/H$, otherwise to an absorbing zero-reward state $w$), and action $B$ pays $1$ and moves to $w$. The initial state is
$x$ and the rewards are known. The design puts mass $0.2$ on each pair at $x$ and the rest on $y$ and $w$ (twice as much on $y$), each split equally
between two identical actions. We used $n=4\cdot10^6$, $\beta=8H$, $H\in\{8,16,32,64\}$ and 120 datasets per $H$, and computed the losses of PSVI with both penalties and of
the plug-in estimator exactly.

\paragraph{The plug-in instance (Theorem~\ref{thm:plugin}).} The instance $I(m,\gamma)$ with $H=20$, $\beta=0.4$ and $m=n/H$ rare actions, $n\in\{2\cdot10^3,2\cdot10^4,2\cdot10^5\}$
(only the largest $n$ satisfies $n\ge2000H$), and 600 datasets per $n$, under (D2$'$) and under (D2). For comparison we also ran PSVI with the
additive penalty $\sqrt{L\widehat\Var/n_z}+HL/n_z$, $L=\log(4Kn)$, which the theorems do not cover (Remark~\ref{rem:additive}). Its mean loss was at most $0.05$
for $n\ge2\cdot10^4$ under both data models; at $n=2\cdot10^3$, below the sample sizes that Theorem~\ref{thm:h4} needs, it reached $\CFivePsviSmallNMax$.

\paragraph{Constructions of the lower bounds.} For Theorems~\ref{thm:lb-design} and~\ref{thm:power} we evaluated the inequalities of the
constructions exactly over grids of $\gamma\in[4/7,0.97]$, of $\beta$ (from $0.01H$ to $10H$, or from $H$ to $5H$ for Theorem~\ref{thm:power}) and of $\Delta$
up to its largest allowed value, with up to ten gadgets or chain states and random policies for the loss identities. For Theorem~\ref{thm:gate}
the grid also covers $C$ from $4$ to the largest value the theorem allows (capped at $10^8$), up to three gates and every sign vector. For
Lemma~\ref{lem:chain} we computed the exponent of the reach probabilities on long chains with $H$ up to $100$ and $\beta/H$ from $1$ to $10$, and compared
it with $\kappa H/\beta$.

\paragraph{Coverage (Theorem~\ref{thm:coverage}, Corollary~\ref{cor:refdata}).} For Proposition~\ref{prop:mgf} we computed the moment $\E_*L^{\beta/(6H)}$ exactly on 900
random MDPs with 2--8 states and $\beta/H$ from $0.02$ to $10$. For Corollary~\ref{cor:refdata} we evaluated, with $E=\{z:d^*(z)>\tau_n\}$, the two steps
that use Theorem~\ref{thm:coverage}: the count condition $n\dref\ge17408H\iota$ on $E$ and $C_\Sigma(E)\le Ke^{c_1H/\beta}\tau_n^{-a}$. The instances were 900 random
MDPs and the chains of Lemma~\ref{lem:chain} with up to $40$ states, at $\beta/H\in\{0.5,1,1.4,2,8\}$ and $n$ from $10^3$ to $10^{24}$. With the power
removed (exponent $1$ in place of $1+a$, the same $c_1$) the count condition fails on long chains (in \CTwentyFourCtrl\ cases); that no bound
without a power can hold follows from Lemma~\ref{lem:chain}.

\paragraph{Lemmas behind the upper bounds.} For Lemma~\ref{lem:pathpess} we used random MDPs with 2--6 states, temperatures from $0.01H$ to $10H$ and
random slacks, together with slacks found by a local search that increases the ratio of the two sides. For Lemmas~\ref{lem:loctilt}--\ref{lem:mixed} and
the energy step of Theorem~\ref{thm:h3}, the pessimistic output is the fixed point of $\hat Q=\max\{0,r+\gamma P\hat V-\zeta_0\}$ with a random nonnegative
$\zeta_0$, so that $0\le\hat Q\le Q^*$, which is all these lemmas use; we drew 1200 such instances at eight temperatures from $1.32H$ to $16H$ for
Lemmas~\ref{lem:loctilt}--\ref{lem:avp}, and 900 at temperatures from $0.1H$ to $16H$ for the energy step and Lemma~\ref{lem:mixed}. The constants of the
$h_-$ tilt bound and of Lemma~\ref{lem:ptv} are close to tight on these instances: halving the first, or replacing $3$ by $1$ in the second, gives
inequalities that some instances violate. The energy
inequalities, on the other hand, are far from tight here: since $g\le H^2/4$, the ratio $\E_s\Gamma/\bar\Gamma$ is at most $H(1-\alpha)^2/32$. For
Proposition~\ref{prop:burnin} we compared the slack with the split of its proof ($c_s=72$) on sampled data at $\beta/H\in\{0.2,1\}$.

\paragraph{The penalty events (Lemma~\ref{lem:contraction}, Propositions~\ref{prop:eventB} and~\ref{prop:subsample}).} For Lemma~\ref{lem:contraction} we drew
one pair's empirical law and a point $V\in[0,H]^{\cS}$, with $H\in\{2,10,100\}$ and the coefficient $\omega=4\sqrt{\iota/n}$ of Appendix~\ref{app:upper} from
$[0.01,0.6]$ (the proof holds for every $\omega$); half of the draws put the next state on two states with values near $0$ and near $H$, so that the
variance term is often active. We compared the exact gradient with finite differences and tested monotonicity and the $\gamma$-Lipschitz
property on hundreds of thousands of pairs $V\le V'$. With the floor halved some partial derivatives are negative. For
Proposition~\ref{prop:eventB} we used random MDPs with small counts, where the variance term is rarely active, and a three-state instance with a
self-loop where it is active at both pairs of the looping state $s_0$. There the two actions move to $s_0$ itself or to one of two absorbing
states, worth $0$ and $1$ per step, and pay Bernoulli rewards; $\beta=H$, and $n$ goes up to $2\cdot10^7$. The fixed point at $s_0$
then depends on the samples at $s_0$, the case the leave-one-out argument is for. For Proposition~\ref{prop:subsample} we ran the subsampled
estimator on reference streams of a random four-state MDP (Hoeffding penalty) and of the self-loop instance (Bernstein penalty). The
deviations stayed well inside the penalties (Table~\ref{tab:checks}), and on these instances the standard-deviation form of
Proposition~\ref{prop:eventB}(ii) is far from tight.

\paragraph{Explicit constants (Appendix~\ref{app:constants}).} We evaluated the bounds of Appendix~\ref{app:constants} on the loop family ($n=4\cdot10^6$) and on the
instance of Theorem~\ref{thm:plugin}, each minimized over all sets $E$ the theorem allows, and compared them with the exact loss of the reference
policy. We found the smallest sample size at which a bound falls below that loss by a scan in $n$ followed by bisection.

\ifneuripschecklist
\clearpage
\input{checklist}
\fi

\end{document}